\documentclass[runningheads]{llncs}
\usepackage{graphicx}

\usepackage{tikz}
\usepackage{comment}
\usepackage{amsmath,amssymb} 
\usepackage{color}
\usepackage{subfigure}
\usepackage{verbatim}
\usepackage{multirow}
\usepackage{enumitem}
\usepackage{mathrsfs}
\usepackage[title]{appendix}
\usepackage[accsupp]{axessibility}  
\usepackage{footmisc}

\usepackage[pagebackref=true,breaklinks=true,colorlinks,bookmarks=false]{hyperref}

\newcommand{\ie}{\textit{i}.\textit{e}.}
\newcommand{\eg}{\textit{e}.\textit{g}.}
\newcommand{\cf}{\textit{cf.}}
\newcommand{\etal}{\textit{et}.\textit{al}.}

\newcommand{\aka}{\textit{a}.\textit{k}.\textit{a}.}

\usepackage{colortbl} 

\definecolor{mygray}{gray}{0.6}
\definecolor{mygray-bg}{gray}{0.9}

\usepackage{pdfrender}
\newcommand*{\boldcheckmark}{%
	\textpdfrender{
		TextRenderingMode=FillStroke,
		LineWidth=.5pt, 
	}{\checkmark}%
}

\begin{document}
\pagestyle{headings}
\mainmatter
\def\ECCVSubNumber{2242}  

\title{Explicit Image Caption Editing} 



\titlerunning{Explicit Image Caption Editing}
%
\author{
    Zhen Wang\inst{1}\thanks{Zhen Wang and Long Chen are co-first authors with equal contributions.} \and 
    Long Chen\inst{2}$^\ast$ \and 
    Wenbo Ma\inst{1} \and 
    Guangxing Han\inst{2} \and 
    Yulei Niu\inst{2} \and \\
    Jian Shao\inst{1} \and 
    Jun Xiao\inst{1}\thanks{Corresponding author. Codes: \url{https://github.com/baaaad/ECE}.}
}
\authorrunning{Z. Wang and L. Chen et al.}
%
\institute{$^1$Zhejiang University \quad
$^2$Columbia University \\
\email{zju\_wangzhen@zju.edu.cn, zjuchenlong@gmail.com, junx@cs.zju.edu.cn}
}

\maketitle

\begin{abstract}
Given an image and a reference caption, the image caption editing task aims to correct the misalignment errors and generate a refined caption. However, all existing caption editing works are \emph{implicit} models, \ie, they directly produce the refined captions without explicit connections to the reference captions. In this paper, we introduce a new task: Explicit Caption Editing (ECE). ECE models explicitly generate a sequence of \emph{edit operations}, and this edit operation sequence can translate the reference caption into a refined one. Compared to the implicit editing, ECE has multiple advantages: 1) Explainable: it can trace the whole editing path. 2) Editing Efficient: it only needs to modify a few words. 3) Human-like: it resembles the way that humans perform caption editing, and tries to keep original sentence structures. To solve this task, we propose the first ECE model: \texttt{TIger}. It is a non-autoregressive transformer-based model, consisting of three modules: Tagger$_{\text{del}}$, Tagger$_{\text{add}}$, and Inserter. Specifically, Tagger$_{\text{del}}$ decides whether each word should be preserved or not, Tagger$_{\text{add}}$ decides where to add new words, and Inserter predicts the specific word for adding. To further facilitate ECE research, we propose two ECE benchmarks by re-organizing two existing datasets, dubbed COCO-EE and Flickr30K-EE, respectively. Extensive ablations on both two benchmarks have demonstrated the effectiveness of \texttt{TIger}.

\keywords{Image Captioning, Caption Editing, Explicit Editing}

\end{abstract}

\section{Introduction} \label{sec:1}

\begin{figure}[t]
	\setlength{\abovecaptionskip}{-1em}
	\setlength{\belowcaptionskip}{-1em}
	\begin{center}
    	\includegraphics[width=\linewidth]{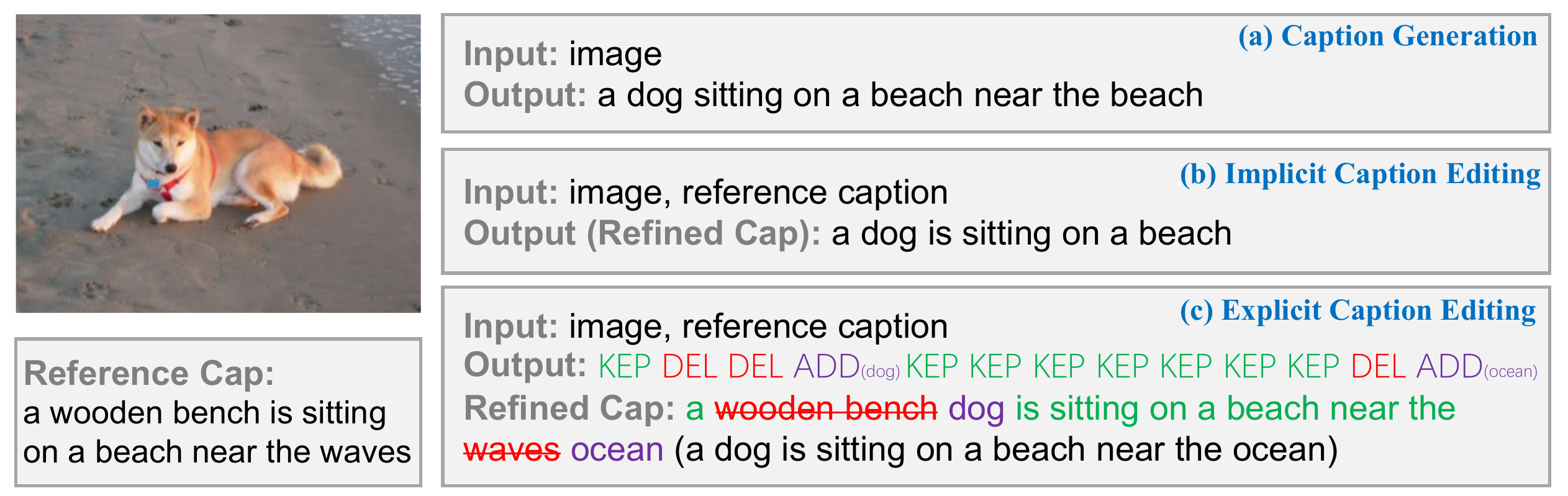}
	\end{center}
	\caption{Comparisons between our proposed ECE task (c) and existing caption generation (a) and implicit caption editing (b). The outputs are from the SOTA models~\cite{anderson2018bottom,sammani2020show}.}
	\label{fig:motivation_comparisons}
\end{figure}

Image caption generation (\aka, image captioning), is the task of generating natural language captions for given images. Due to its multimodal nature and numerous downstream applications (\eg, human-machine interaction~\cite{das2017visual}, content-based image retrieval~\cite{ordonez2016large}, and assisting visually-impaired people~\cite{macleod2017understanding}), caption generation has raised unprecedented attention from both CV and NLP communities. Thanks to the development of encoder-decoder frameworks (\eg, CNN+ RNN~\cite{vinyals2015show} or Transformer~\cite{vaswani2017attention}), current state-of-the-art image caption generation models can generate ``reasonable" captions from scratch and achieve satisfactory performance. However, numerous studies~\cite{sammani2019look,sammani2020show} have revealed that these SOTA models always suffer from severe bias issues and overlook some content details (\eg, gender bias~\cite{hendricks2018women}, object hallucination~\cite{rohrbach2018object}). As shown in Fig.~\ref{fig:motivation_comparisons}(a), given the input image, a SOTA captioning model~\cite{anderson2018bottom} generates ``\texttt{a dog sitting on a beach near the beach}". Thus, SOTA models can indeed generate a coherent sentence structure for the image (\ie, ``\texttt{a \_\_ on a \_\_ near the \_\_}"), but fail to properly predict the correct details and even repeat the main object ``\texttt{beach}".

To mitigate these problems and make the generated captions focus more on visually-grounded content details (beyond sentence structures), some pioneering works~\cite{sammani2019look,sammani2020show} have proposed a new task: Image Caption Editing (ICE). Different from captioning models which generate captions from scratch, ICE directly edits another reference caption and pays more attention to the misaligned details. For example in Fig.~\ref{fig:motivation_comparisons}(b), ICE model takes an extra reference caption ``\texttt{a wooden bench is sitting on a beach near the waves}" as input, and aims to generate a refined caption. Unfortunately, all existing ICE works are \textbf{\emph{implicit}} editing models. By ``implicit", we mean that they directly produce final refined captions, without explicit connections (editing process) to the reference captions.

Although ICE models can significantly improve the captions qualities, it is worth noting that there are still several drawbacks for this implicit manner: 1) \textbf{Unexplainable}: they fail to explain whether these words are copied from the reference caption or regenerated, and whether they truly recognize and modify errors or simply generate words by language priors~\cite{lu2017knowing}. 2) \textbf{Inefficient}: All words are regenerated, which is more like rewriting or re-captioning instead of editing. 3) \textbf{Structure-breaking}: They are easy to break the sentence structures of reference captions without focusing on details. For example in Fig.~\ref{fig:motivation_comparisons}(b), the model roughly deletes part of the structure (\eg, ``\texttt{near the \_\_}").

In this paper, we introduce a new image caption editing task: \textbf{Explicit Caption Editing} (ECE). By ``explicit", we mean that ECE models explicitly generate a sequence of \emph{edit operations}, and these edit operations translate the reference captions into the refined captions. Typically, the edit operations consist of \texttt{ADD}, \texttt{DELETE}, and \texttt{KEEP}\footnote{These are the most common edit operations in numerous text explicit editing tasks, such as simplification~\cite{dong2019editnts,malmi2019encode}, fusion~\cite{mallinson2020felix}. Of course, different ECE models can design or propose other edit operations, \eg, \texttt{REORDER}. More discussion are left in appendix.}. As shown in Fig.~\ref{fig:motivation_comparisons}(c), for each input word in the reference caption, the ECE model predicts \texttt{KEEP} or \texttt{DELETE} to decide whether this word needs to be preserved or not, and predicts \texttt{ADD} to add extra specific words. The predicted edit operation sequence is mainly composed with \texttt{KEEP} to preserve the main sentence structure and few \texttt{DELETE}/\texttt{ADD} to fix misalignment errors. Compared to existing implicit caption editing works, ECE avoids all mentioned weaknesses: 1) ECE traces the whole editing path, which is used to translate reference captions (\textbf{Explainable}). 2) ECE only needs to modify a few words (\textbf{Explicit Editing Efficient}). 3) ECE resembles the way that humans perform editing, and tries to keep the original sentence structures (\textbf{Structure-preserving}).

To solve this new task, we propose the first ECE model, a non-autoregressive transformer-based ECE model: \texttt{TIger} (\underline{\textbf{T}}ag\underline{\textbf{ger}} and \underline{\textbf{I}}nserter). Specifically, \texttt{TIger} consists of three modules: Tagger$_{\text{del}}$, Tagger$_{\text{add}}$, and Inserter. All three modules are built on top of the multimodal BERT architecture~\cite{lu2019vilbert}. Given an input image and a reference caption, Tagger$_{\text{del}}$ decides whether each word should be preserved or not by predicting \texttt{KEEP} and \texttt{DELETE}. Then, Tagger$_{\text{add}}$ decides whether a new word should be added after each input word by predicting \texttt{KEEP} and \texttt{ADD}. A special token [\texttt{Mask}] is placed for each position with the \texttt{ADD} prediction. Subsequently, Inserter predicts the specific word for each [\texttt{Mask}] token. Since Tagger$_{\text{add}}$ only adds one new word after each input word once a time, we iteratively execute Tagger$_{\text{add}}$ and Inserter multiple rounds to guarantee enough words adding.

To further facilitate ECE research, we also propose two new ECE benchmarks by re-organizing MSCOCO~\cite{lin2014microsoft} and e-SNLI-VE~\cite{young2014image,kayser2021vil}, dubbed \textbf{COCO-EE} and \textbf{Flickr30K-EE}, respectively. Particularly, we pair each reference caption with one ground-truth caption by several criteria and rules. Each ECE instance consists of an image, a reference caption, and a ground-truth caption. Compared to existing implicit editing works~\cite{sammani2020show,sammani2019look} which use machine-generated captions as reference captions, ours are all human-written sentences, \ie, they are more natural and have no grammatical errors. Besides, we propose two supplementary metrics for ECE: Editing Steps (ES) and Gains Per Step (GPS), which consider not only the quality of captions, but also the efficiency of editing models.

In summary, we make three main contributions:
1) We propose a new visual-language task: ECE, \ie, the caption editing model explicitly generates a set of edit operations on the reference captions.
2) For reliable benchmarking, we propose two new ECE datasets (COCO-EE and Flickr30K-EE), and new metrics for ECE evaluation.
3) We propose the first ECE model \texttt{TIger}. Extensive ablations have demonstrated the effectiveness of \texttt{TIger}. Moreover, \texttt{TIger} can serve as an off-the-shelf model to improve the quality of machine-generated captions.

\section{Related Work}

\noindent\textbf{Image Caption Generation.} With the release of advanced encoder-decoder frameworks, NN-based~\cite{mao2014deep,karpathy2015deep,vinyals2015show} methods have risen to prominence. They typically use an encoder to extract image features and a decoder to generate all words. Recent advances in captioning works focus on stronger architectures and better training procedures. To encoder visual context, numerous attention mechanisms are proposed to boost the performance~\cite{xu2015show,chen2017sca,yang2016stacked,anderson2018bottom,huang2019attention,mao2022rethinking,liu2021region,wang2021high}, and they tend to focus on specific local features in the image when predicting each word in the caption. On the other side, current caption generation performance is dominated by reinforcement learning (RL) based methods~\cite{ren2017deep,rennie2017self,xu2019multi}, which directly optimize the sequence-level caption quality. Besides, to accelerate the decoding process, non-autoregressive methods~\cite{fei2019fast,gao2019masked,guo2020non} are proposed, which simultaneously generate words by discarding the sequential dependencies within sentence.

\begin{figure*}[t]
    \setlength{\abovecaptionskip}{-1em}
    \setlength{\belowcaptionskip}{-1em}
    \begin{center}
        \includegraphics[width=\linewidth]{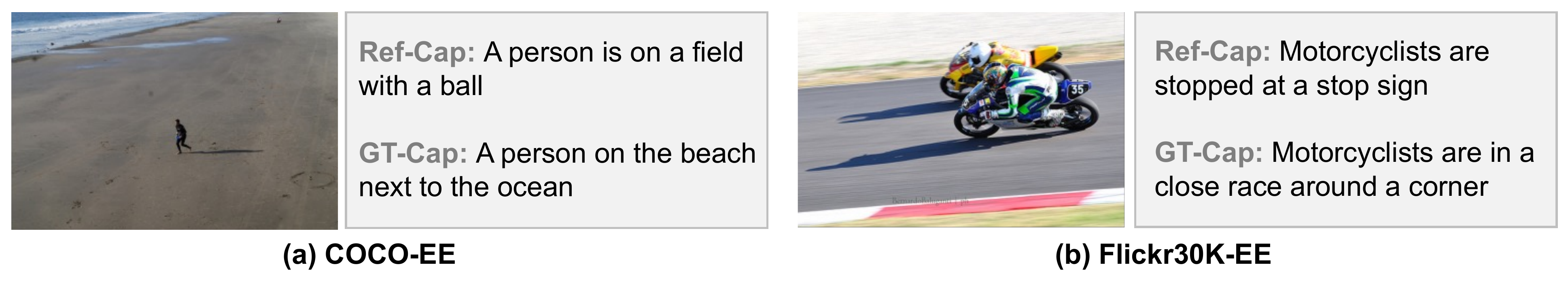}
    \end{center}
    \caption{Two examples from proposed ECE benchmarks: COCO-EE and Flickr30K-EE.}
    \label{fig:example_COCOEE_Flickr30KEE}
\end{figure*}

\noindent\textbf{Image Caption Editing.}
ICE, \ie, editing the existing reference caption paired with an image for refinement instead of re-generating from scratch, was first proposed by Sammani~\etal~\cite{sammani2019look}. Specifically, they use a pre-trained deep averaging network to encode the reference caption, and design a gate mechanism to help the decoder to generate refined captions. Later, Sammani~\etal~\cite{sammani2020show} proposed a new method for caption editing, which designs a selective copy memory attention to better encode the reference caption. As discussed above, they are all \emph{implicit} caption editing models. In this paper, we propose the new explicit editing task, which can avoid the weaknesses in existing implicit works.

\noindent\textbf{Explicit Text Editing.}
Explicit text editing, explicitly labeling the input reference caption with a sequence of edit operations, has been widely applied in different text editing tasks, such as text simplification~\cite{alva2017learning,dong2019editnts}, sentence fusion~\cite{malmi2019encode,mallinson2020felix}, grammatical error correction~\cite{awasthi2019parallel} and text generation~\cite{gu2019levenshtein}. Besides the basic edit operations like insertion and deletion, they tend to design different edit operations and edit mechanisms for their specific downstream tasks. In this paper, we extend three explicit text editing models (EditNTS~\cite{dong2019editnts}, LaserTagger~\cite{malmi2019encode}, and Felix~\cite{mallinson2020felix}) into ECE, and compare them with our \texttt{TIger}. Specifically, EditNTS predicts edit operations by an LSTM sequentially. LaserTagger and Felix are all Transformer-based models, where LaserTagger predicts the edit operations restricted to a fixed phrase vocabulary and Felix uses extra reordering operations.

\section{ECE and Benchmarks}

\subsection{Task Definition: Explicit Caption Editing (ECE)}

In this section, we first formally define the ECE task. Given an image and a reference caption (Ref-Cap), ECE models aim to explicitly predict a sequence of edit operations (\eg, \texttt{KEEP}/\texttt{DELETE}/\texttt{ADD}) on the Ref-Cap, which can translate the Ref-Cap close to the ground-truth caption (GT-Cap). Typically, Ref-Cap is slightly misaligned with the image. This task hopes the captioning models not only focus more on the visually-grounded content details, but also perform more explainable, explicit editing efficient\footnote{We emphasize efficient from the perspective of ``explicit editing efficiency", as realizing more performance gains with less meaningful editing steps, which differs from
other efficiency metrics (inference time and FLOPs). More details are left in appendix.}, and human-like editing. As the example shown in Fig.~\ref{fig:example_COCOEE_Flickr30KEE}(b), given Ref-Cap ``\texttt{Motorcyclists are stopped at a stop sign}", the ECE models aim to explicitly predict a edit operation sequence: ``\texttt{KEEP}$_{\texttt{Motorcyclists}}$ \texttt{KEEP}$_{\texttt{are}}$ \texttt{DELETE}$_{\texttt{stopped}}$ \texttt{DELETE}$_{\texttt{at}}$ \texttt{ADD}$_{\texttt{in}}$ \texttt{KEEP}$_{\texttt{a}}$ \texttt{DELETE}$_{\texttt{stop}}$ \texttt{DELETE}$_{\texttt{sigh}}$ \texttt{ADD}$_{\texttt{close}}$ \texttt{ADD}$_{\texttt{race}}$ \texttt{ADD}$_{\texttt{around}}$ \texttt{ADD}$_{\texttt{a}}$ \texttt{ADD}$_{\texttt{corner}}$"\footnote{Based on different basic edit operations used in each ECE model, the GT edit operation sequence can be different. This example uses \texttt{KEEP}/\texttt{DELETE}/\texttt{ADD} as operations.}. 

\subsection{Explicit Caption Editing Benchmarks} \label{sec:3.2}

\noindent\textbf{Criteria.} Based on the task definition of ECE and essential requirements of each ECE instance, each reference caption (Ref-Cap) and its corresponding ground-truth caption (GT-Cap) should be selected reasonably for each image. We argue that there are several criteria in developing high-quality ECE datasets:

\emph{c1.} \textbf{Human Annotated Captions.} Both Ref-Cap and GT-Cap should be written by humans to avoid grammatical errors.

\emph{c2.} \textbf{Image-Caption Similarity.} The scene described by the Ref-Cap should be similar to the scene in the image.
    
\emph{c3.} \textbf{Caption Similarity.} Paired Ref-Cap and GT-Cap should have a certain degree of overlap and similar caption structure to avoid completely regenerating the whole sentence or roughly breaking the structure of Ref-Cap.
    
\emph{c4.} \textbf{Caption Differences.} To ensure necessary editing operations, the differences between the Ref-Cap and GT-Cap shouldn't be just one (or few) words, which can be easily corrected by only language bias.

Existing ICE work~\cite{sammani2020show,sammani2019look} simply uses machine-generated captions as their Ref-Caps, which may mislead editing models to focus more on grammatical errors instead of content details. Meanwhile, each image has five GT-Caps, and these GT-Caps may have potential differences (caption structures or described events~\cite{chen2021human}). These training samples may confuse the editing model to break the sentence structures of Ref-Caps. To this end, we constructed two high-quality ECE benchmarks based on the aforementioned criteria. Details are as follows:

\textbf{COCO-EE.}
We built COCO-EE based on dataset MSCOCO~\cite{lin2014microsoft}, which contains 123,287 images, and 5 ground-truth captions for each image. To ensure \emph{c1}, we selected all Ref-Caps and GT-Caps in COCO-EE from MSCOCO captions. Since each image is labeled with 5 captions, we regard all 5 ground-truth captions as the GT-Cap candidates and filter Ref-Cap candidates from the rest captions based on image-caption similarity score to ensure \emph{c2}. We then calculated several caption similarity scores to further filter the Ref-Cap candidates to ensure \emph{c3} and \emph{c4}. Finally, for each filtered Ref-Caps candidate, we selected the caption with the shortest edit distance\footnote{The shortest edit distance is the minimum number of edit operations (except the \texttt{KEEP} operation) to translate one sentence to the target sentence.} from corresponding GT-Caps candidates to form a Ref-GT caption pair. Following the above steps\footnote{More details about the dataset construction steps are left in the appendix. \label{foot:dataset}}, we constructed COCO-EE, and divided it into training, val, and test sets following the “Karpathy” split~\cite{karpathy2015deep}. The statistical summary about COCO-EE is shown in Table~\ref{tab:dataset}.

\begin{table}[t]
  \setlength{\abovecaptionskip}{-0.5em}
  \setlength{\belowcaptionskip}{-1em}
  \begin{center}
    \begin{tabular}{l|ccc|ccc}
    \hline
    \multirow{2}{*}{Dataset} & \multicolumn{3}{c|}{\textbf{COCO-EE}} & \multicolumn{3}{c}{\textbf{Flickr30K-EE}}\\
    & {Train} & {Val} & {Test} & {Train} & {Val} & {Test}\\
    \hline
    {\#Editing instances} & {97,567} & {5,628} & {5,366} & {108,238} & {4,898} & {4,910}\\
    {\#Images} & {52,587} & {3,055} & {2,948} & {29,783} & {1,000} & {1,000}\\
    {Mean Reference Caption Length} & {10.3} & {10.2} & {10.1} & {7.3} & {7.4} & {7.4}\\
    {Mean Ground-Truth Caption Length} & {9.7} & {9.8} & {9.8} & {6.2} & {6.3} & {6.3}\\
    {Mean Edit Distance} & {10.9} & {11.0} & {10.9} & {8.8} & {8.8} & {8.9}\\
    {Vocabulary} & {11,802} & {3,127} & {3,066} & {19,124} & {4,178} & {4,183}\\
    \hline
    \end{tabular}%
  \end{center}
  \caption{Statistical summary of the COCO-EE and Flickr30K-EE benchmarks.}
  \label{tab:dataset}%
\end{table}%

\textbf{Flickr30K-EE.} We built Flickr30K-EE based on dataset e-SNLI-VE~\cite{kayser2021vil}. e-SNLI-VE is a visual entailment dataset using the same image set as the image captioning dataset Flicrk30K~\cite{young2014image}. For each image in e-SNLI-VE, there are three sentences (hypothesis), which have different relations with the image (premise): entailment, neutral, and contradiction. For each image and its textual hypotheses in e-SNLI-VE, we selected the contradiction and entailment hypothesis as a Ref-GT caption pair if they have the same text premise, which ensures \emph{c2}. Since the paired contradiction and entailment hypothesis are human-annotated (\emph{c1}) and have the same text premises, they tend to have a certain textual similarity (\emph{c3}) while maintaining visual differences (\emph{c4}) at the same time. Together with the image, each ECE instance contains one image from Flickr30K and one human-annotated Ref-Cap and GT-Cap pair. Finally, we obtained the Flickr30K-EE\footref{foot:dataset}. Similarly, we divided it into training, val, and test sets based on e-SNLI-VE splits. The statistical summary about Flickr30K-EE is shown in Table~\ref{tab:dataset}.

\section{Proposed Approach}

\begin{figure}[t]
	\setlength{\abovecaptionskip}{-1em}
    \begin{center}
    	\includegraphics[width=\linewidth]{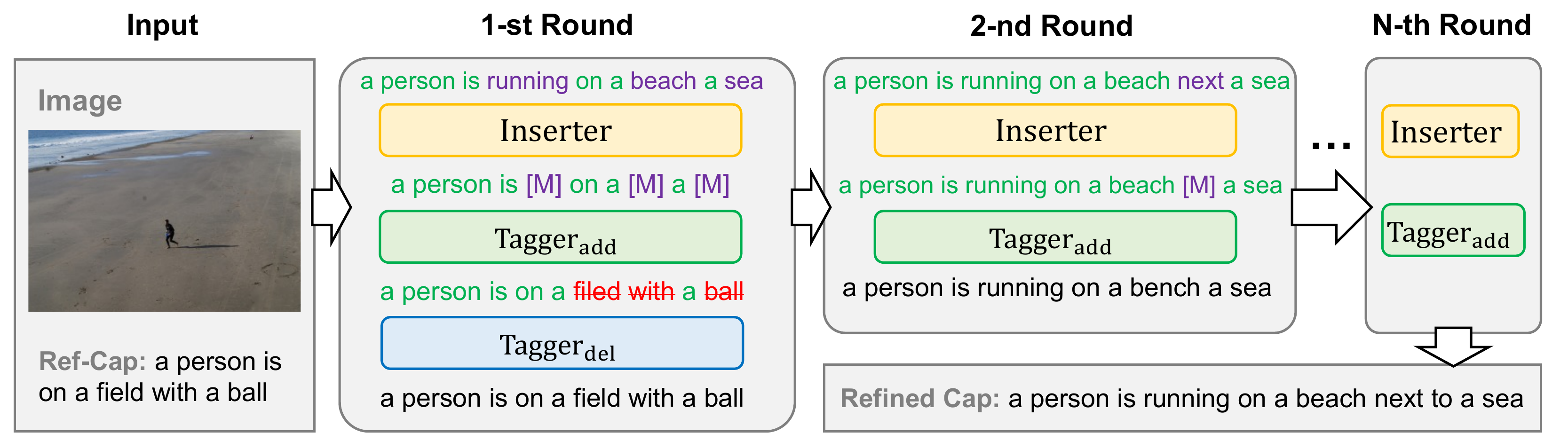}
    \end{center}
	\caption{Overview of the whole \texttt{TIger} pipeline. Tagger$_{\text{del}}$ is only used in the first round, Tagger$_{\text{add}}$ and Inserter are used in all rounds. In the first editing round, \texttt{TIger} aims to fix the main errors. Then, in the following rounds, \texttt{TIger} tries to add more details to generate more coherent and reasonable captions. [M] denotes the special [\texttt{MASK}] token.}
	\label{fig:multitime_editing}
\end{figure}

\noindent\textbf{Overview.}
In this section, we introduce the proposed \texttt{TIger} for the ECE task. Specifically, the design of the \texttt{TIger} is inspired from the manner in which humans conduct caption editing, \ie, \emph{our humans would like to delete all the irrelevant or wrong words in the reference caption first, and then gradually add the missing words or details till enough.} Based on this motivation, we design three modules in \texttt{TIger}: \textbf{Tagger$_{\text{del}}$}, \textbf{Tagger$_{\text{add}}$}, and \textbf{Inserter}. The overview of the pipeline of the \texttt{TIger} is illustrated in Fig.~\ref{fig:multitime_editing}, and the function of each module is as follows:

\textbf{1) Tagger$_{\text{del}}$}: The Tagger$_{\text{del}}$ aims to predict whether to keep or delete each input word. For example in Fig.~\ref{fig:multitime_editing} (1-st Round), the words ``\texttt{field}", ``\texttt{with}" and ``\texttt{ball}" in the reference caption (``\texttt{a person is on a \underline{filed} \underline{with} a \underline{ball}}") are not related to the image content, and we hope the Tagger$_{\text{del}}$ module can predict ``\texttt{DELETE}" for these words, and ``\texttt{KEEP}" for the rest of the words.

\textbf{2) Tagger$_{\text{add}}$}: The Tagger$_{\text{add}}$ aims to decide which words need to be added with a new word after them, and a special token  [\texttt{Mask}] will be placed after these words. For example, given the input caption (``\texttt{a person is on a a}"), Tagger$_{\text{add}}$ thinks a new word should be added after ``\texttt{is}", ``\texttt{a}", and ``\texttt{a}", \ie, the output of Tagger$_{\text{add}}$ is ``\texttt{a person is [\texttt{Mask}] on a [\texttt{Mask}] a [\texttt{Mask}]}".

\textbf{3) Inserter}: Given the output of Tagger$_{\text{add}}$, the Inserter aims to predict a specific word for each [\texttt{Mask}] token, \ie, ``\texttt{running}", ``\texttt{beach}", and ``\texttt{sea}".

Since the Tagger$_\text{add}$ and Inserter can only add one new word at each position for each round, we can easily run Tagger$_\text{add}$ and Inserter iteratively for multiple rounds to guarantee enough words adding. Instead, for the Tagger$_\text{del}$, we hope it directly detects all the wrong or unsuitable words in the first round.

\subsection{Multimodal Feature Extraction}
As shown in Fig.~\ref{fig:method}, all three modules Tagger$_{\text{del}}$,  Tagger$_{\text{add}}$, and Inserter are all built on top of the multi-modal BERT~\cite{lu2019vilbert,li2020unicoder}, which applies a series of transformer blocks and co-attention layers to learn better multi-modal features of the images and texts. The input for each module is a sequence of multimodal tokens.

\noindent\textbf{Visual Token Representations.} For the given image, we first generate a set of image region features by extracting proposals and their corresponding visual features from a pre-trained object detector. We also encode the spatial location features of each proposal into a 5-d vector (normalized top-left and bottom-right coordinates, and fraction of the region area covered). A visual token feature is the sum of a region proposal feature and its spatial location feature. In addition, a special [\texttt{IMG}] token is placed at the beginning of the visual token sequence to represent the entire image. The token feature of [\texttt{IMG}] is the mean-pooled visual feature with a spatial encoding corresponding to the entire image.

\noindent\textbf{Textual Token Representations.} For the given reference caption, we first convert it into a sequence of tokens by tokenization~\cite{devlin2018bert}. Then, we put special [\texttt{CLS}] and [\texttt{SEP}] tokens at the start and end of textual token sequence, respectively. Meanwhile, for Inserter, another token [\texttt{MASK}] is used to indicate the position for new words adding. Same as~\cite{lu2019vilbert}, a textual token representation is the sum of token-specific learned embedding~\cite{wu2016google}, position encoding, and segment encoding.

\noindent\textbf{Multimodal Input Token Sequence.} Given the image and reference caption, we first encoder them into a sequence of visual tokens $\{v_1, \dots,v_K\}$ and textual tokens \{$w_1, \dots,w_L\}$, respectively. $K$ and $L$ is the number of visual and textual tokens, respectively. Then, the input token sequence for the three modules is $\{[\texttt{IMG}], v_1, \dots, v_K, [\texttt{CLS}], w_1, \dots, w_L, [\texttt{SEP}]\}$. The output representations for the visual and textual tokens are $\{h_{v_1}, \dots, h_{v_K}\}$ and $\{h_{w_1}, \dots, h_{w_L}\}$, respectively.

\subsection{Model Description}
\begin{figure}[t]
	\setlength{\abovecaptionskip}{-1em}
	\setlength{\belowcaptionskip}{-1em}
    \begin{center}
    	\includegraphics[width=0.98\linewidth]{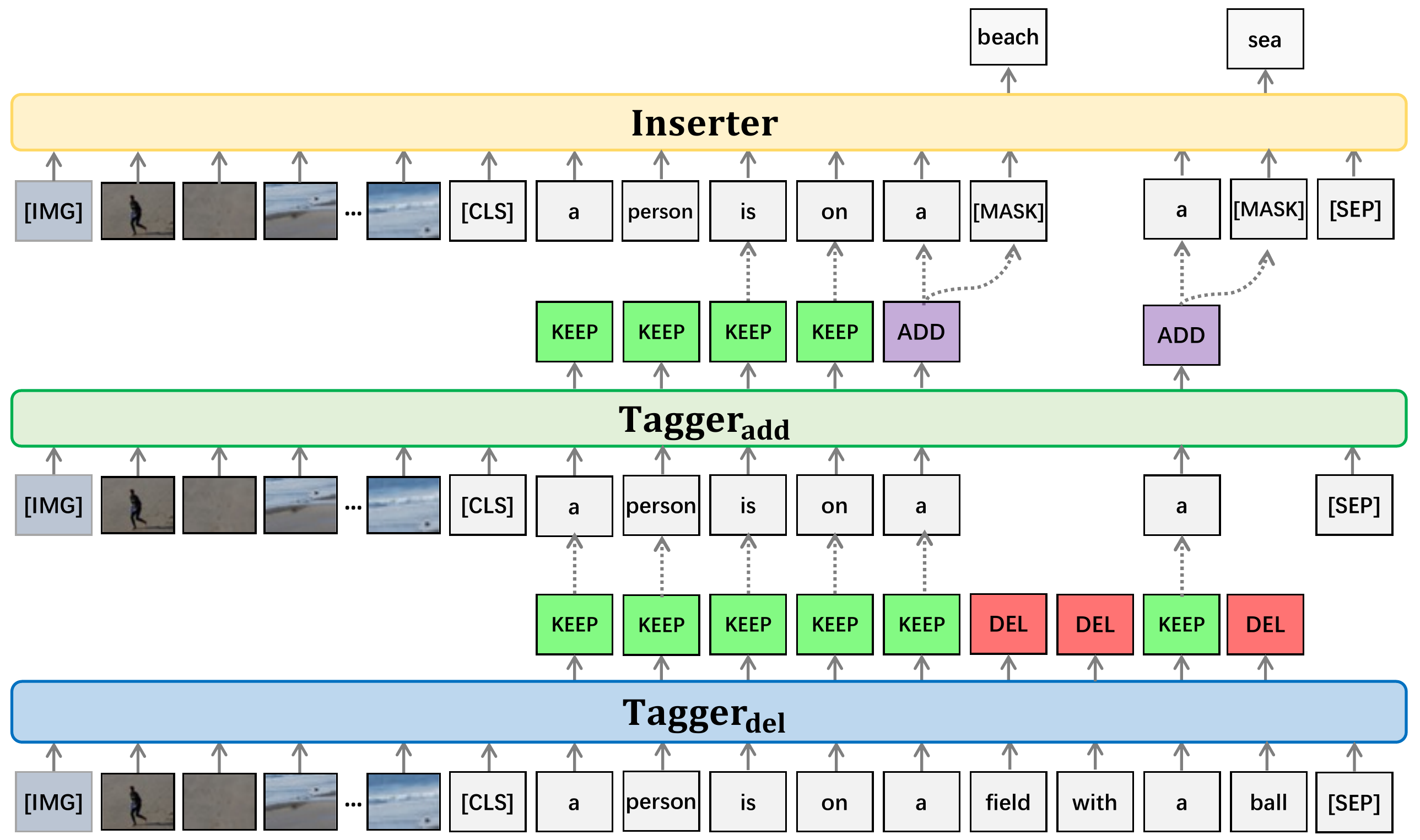}
    \end{center}
	\caption{Illustration of the input visual-language token sequences for each module. We take the first editing rounds in as the example.}
	\label{fig:method}
\end{figure}

\noindent\textbf{Tagger$_\text{del}$ \& Tagger$_\text{add}$ Modules.}
As shown in Fig.~\ref{fig:method}, given the visual-textual token sequence, Tagger$_{\text{del}}$ and Tagger$_{\text{add}}$ tag each textual token with a specific edit operation $z$. For each textual token, both Tagger$_{\text{del}}$ and Tagger$_{\text{add}}$ conduct a binary classification, \ie, $z\in \{ \texttt{KEEP},\texttt{DELETE} \}$ for Tagger$_{\text{del}}$ and $z\in \{ \texttt{KEEP},\texttt{ADD} \}$ for Tagger$_{\text{add}}$. We pass the final representation of each textual token $\{h_{w_1}, h_{w_2}, \dots, h_{w_L}\}$ into a two-layer MLP to make the binary prediction, \ie, $z_{w_i}= \mathop{\arg\max}f(h_{w_i})$. Thus, the entire output of Tagger$_{\text{del}}$ and Tagger$_{\text{add}}$ is a sequence of edit operations corresponding to the sequence of input tokens, represented as $\{z_{w_1}, z_{w_2}, \dots, z_{w_L}\}$. The output textual token sequence can be translated from the input textual token sequence and predicted edit operations.

\noindent\textbf{Inserter Module.}
As shown in Fig.~\ref{fig:method}, the input tokens fed into the Inserter is a sequence of tokens including the word tokens and the [\texttt{MASK}] tokens, which is constructed from the Tagger$_{\text{add}}$ module. Given the image and the input tokens, the Inserter finishes the insertion by predicting the specific word from the vocabulary for each [\texttt{MASK}] token based on the observed tokens and visual information. Specifically, we pass the final representation of each [\texttt{MASK}] token $h_{w_{mask}}$ into a linear layer, mapping it to a distribution over the vocabulary. Lastly, all [\texttt{MASK}] tokens can be replaced with the predicted word, and the output textual token sequence can be formed with the rest word tokens for following the procedures.

\noindent\textbf{Multi-Rounds Editing.}
As the specific editing process shown in Fig.~\ref{fig:method}, \texttt{TIger} resembles the way that humans might perform caption editing, \ie, considering what to keep, where to add, and what to add. By tracing these edit operations, the whole editing process is explainable and efficient.
Meanwhile, since Tagger$_{\text{add}}$ only adds one new word after each input word once a time, there might not be enough details if we only apply Tagger$_{\text{add}}$ once. Thanks to this modular design, we can seamlessly use Tagger$_{\text{add}}$ and Inserter iteratively for multi-rounds to guarantee enough details. Instead, if we make the Tagger$_{\text{add}}$ can add more than one word once a time, it also needs to predict the number of new words to add at the same time. Meanwhile, the Inserter needs to predict words for multiple [\texttt{MASK}] tokens that may be placed consecutively. This significantly increases the difficulty of training, and empirically this single-round solution gets worse results.

\subsection{Training Objectives} \label{sec:4.3}
The Tagger$_{\text{del}}$ and Tagger$_{\text{add}}$ are essentially solving a binary classification task, and the Inserter is essentially solving a masked language modeling task. Thus, we train all three modules with the cross-entropy (XE) loss. Due to the modular nature, we train the three modules separately. In our experiments, we also emphasize the importance of predicting relative more \texttt{KEEP} operation. Specifically, for Tagger$_{\text{del}}$, it can preserve more words in the caption for the whole following editing process. For Tagger$_{\text{add}}$, it can offer more context words with relative fewer [\texttt{MASK}] tokens for Inserter, which makes the edit operation prediction much easier. Thus, We use different XE loss weights for the \texttt{KEEP} tokens and other tokens (\texttt{DELETE} or \texttt{ADD}). The loss weight ratio $\lambda$ denotes the XE loss weights of edit token \texttt{KEEP/DELETE} for training Tagger$_{\text{del}}$ and \texttt{KEEP/ADD} for training Tagger$_{\text{add}}$, respectively. More detailed influence of $\lambda$ is discussed in Sec.~\ref{sec:5.3}.

\section{Experiments}

\subsection{Experimental Setup}
\label{sec:5.1}
\noindent\textbf{Evaluation Datasets and Metrics.} 
We evaluated our \texttt{TIger} on both COCO-EE and Flickr30K-EE datasets (cf. Sec.~\ref{sec:3.2}). For the caption quality evaluation, we followed existing caption generation works, and used four prevalent evaluation metrics: BLEU-N (B-N) (1-to 4-grams)~\cite{papineni2002bleu}, ROUGE-L (R)~\cite{lin2004rouge}, CIDEr-D (C)~\cite{vedantam2015cider} and SPICE (S)~\cite{anderson2016spice}. Particularly, we evaluated generated captions against its single ground-truth caption. Meanwhile, to evaluate the explicit editing efficiency of editing, we propose two supplementary metrics: Editing Steps (\textbf{ES}), and Gains Per Step (\textbf{GPS}). ES is the total number of meaningful editing steps, and GPS is the average performance gains per meaningful editing step, \ie, we hope ECE models realize the most performance gains with the least number of meaningful editing steps. In this paper, since all baselines apply the same set of edit operations (\ie, \texttt{KEEP}, \texttt{DELETE}, and \texttt{ADD}), we regard the sum of \texttt{DELETE} and \texttt{ADD} operations as ES. Meanwhile, since CIDEr-D is regarded as the most important metric for caption evaluation as to its high agreements with humans, we use the improvements of CIDEr-D score to calculate GPS, denoted as GPS(C).

\noindent\textbf{Baselines.} 
We compared our \texttt{TIger} against state-of-the-art image caption editing models. Specifically, we compared three strong implicit caption editing models: \textbf{UpDn-E}~\cite{anderson2018bottom}, \textbf{MN}~\cite{sammani2019look}, and \textbf{ETN}~\cite{sammani2020show}. They are all built on top of the widely-used UpDn architecture~\cite{anderson2018bottom}, and propose some extra modules to encode the reference caption. Meanwhile, for more complete comparisons, we further extended three text explicit editing models (EditNTS~\cite{dong2019editnts}, LaserTagger~\cite{malmi2019encode}, and Felix~\cite{mallinson2020felix}) into ECE, denoted as \textbf{V-EditNTS}, \textbf{V-LaserTagger}, and \textbf{V-Felix}, respectively. For all these three models, their basic editing operations are \texttt{KEEP}, \texttt{DELETE} and \texttt{ADD}. Specifically, V-EditNTS predicts the edit operation sequence iteratively by an LSTM. V-LaserTagger and V-Felix are one-round Transformer-based editing models, which directly predict multiple \texttt{ADD} operations simultaneously. More details about these baselines are left in the appendix.

\noindent\textbf{Implementation Details.} The implementation details are left in appendix.

\begin{table}[t]
  \setlength{\abovecaptionskip}{-0.5em}
  \setlength{\belowcaptionskip}{-1em}
  \begin{center}
    \scalebox{0.95}{
    \begin{tabular}{c|l|ccccccc|cc|cc}
    \hline
    & \multirow{2}{*}{Model} & \multicolumn{7}{c|}{Quality Evaluation} &
    \multicolumn{4}{c}{Efficiency Evaluation}\\
    & & {B-1} & {B-2} & {B-3} & {B-4} & {R} & {C} & {S} & 
    {ES} & {GPS(C)} & \textcolor{gray}{D} & \textcolor{gray}{A}\\
    \hline
    & Ref-Caps & 50.0  & 37.1  & 27.7  & 19.5  & 48.2  & 129.9  & 18.9  & 
    {---} & \cellcolor{mygray-bg}{---} & {---} & {---} \\
    & UpDn~\cite{anderson2018bottom} & 49.9  & 35.3  & 25.5  & 18.8  & 48.3  & 159.2  & 31.2 & {---} & \cellcolor{mygray-bg}{---} & {---} & {---}  \\
    \hline
    \multirow{3}{*}{ICE} & UpDn-E~\cite{anderson2018bottom} & 54.0  & 40.1  & 30.2  & 22.9  & 52.8  & 182.0  & 33.2 & \textcolor{gray}{19.22}  & \cellcolor{mygray-bg}{\textcolor{gray}{2.71}} & \textcolor{gray}{10.14} & \textcolor{gray}{9.08}\\
    & MN~\cite{sammani2019look} & 50.2  & 35.8  & 26.0  & 19.4  & 48.9  & 163.9  & 31.6 & \textcolor{gray}{19.08}  & \cellcolor{mygray-bg}{\textcolor{gray}{1.78}} & \textcolor{gray}{10.14} & \textcolor{gray}{8.94}\\
    & ETN~\cite{sammani2020show} & 53.8  & 40.5  & 23.8  & 23.8  & 53.3  & 190.5  & 32.1 
    & \textcolor{gray}{18.96} & \cellcolor{mygray-bg}{\textcolor{gray}{3.20}} & \textcolor{gray}{10.14} & \textcolor{gray}{8.82}\\
    \hline
    \multirow{4}{*}{ECE} & V-EditNTS~\cite{dong2019editnts} & 49.2  & 36.5  & 27.4  & 20.5  & 49.8  & 149.0  & 26.2 & 5.90 & \cellcolor{mygray-bg}{3.24} & \textcolor{gray}{3.76} & \textcolor{gray}{2.14} \\
    & V-Felix~\cite{mallinson2020felix} & 36.9  & 28.2  & 21.6  & 16.2  & 49.7  & 139.5  & 25.3 & 5.51 & \cellcolor{mygray-bg}{1.74} & \textcolor{gray}{4.57} & \textcolor{gray}{0.94}\\
    & V-LaserTagger~\cite{malmi2019encode} & 42.0  & 30.5  & 22.4  & 16.0  & 46.8  & 127.1  & 24.1 & 4.11 & \cellcolor{mygray-bg}{-0.68} & \textcolor{gray}{3.54} & \textcolor{gray}{0.57}\\
    \cline{2-13}
    & \textbf{TIger (Ours)} & \textbf{54.8} & \textbf{42.0} & \textbf{32.4} & \textbf{24.7} & \textbf{54.3} & \textbf{194.8} & \textbf{33.3} & 7.74 & \cellcolor{mygray-bg}{\textbf{8.38}} & \textcolor{gray}{4.59} & \textcolor{gray}{3.15}\\
    \hline
    \end{tabular}%
  }
  \end{center}
  \caption{Performance of our model and other state-of-art models on COCO-EE. ``Ref-Caps" denotes the quality of given reference captions. ``D" and ``A" denotes the number of editing step of \texttt{DELETE} and \texttt{ADD} operations, respectively.}
  \label{tab:cocoee}%
\end{table}%

\subsection{Comparisons with State-of-the-Arts}
\label{sec:5.2}
\noindent\textbf{Settings.}
We evaluated \texttt{TIger} on COCO-EE and Flickr30K-EE by comparing with state-of-the-art methods. Since our target is to propose the ECE task and the first ECE model, we first compared \texttt{TIger} with simple ECE baselines which were extended by text explicit editing models (V-EditNTS, V-Felix, and V-LaserTagger). For completeness, we also reported the results of all existing ICE models (UpDn-E, ETN, and MN). Since all implicit models are built on top of the widely-used UpDn architecture, we only reported the results of the UpDn captioning model rather than all other SOTA captioning models (e.g., VLP\cite{zhou2020unified}) as they actually don’t belong to the caption editing task. Since V-Felix and V-LaserTagger are also Transformer-based architectures, we used the same ViLBERT pretrained weights as \texttt{TIger}. For the other baselines, we converted all the words in each dataset to lower cases and built their respective vocabulary. All baselines were trained with XE loss. Since implicit models do not explicitly predict edit operations, we suppose they delete all the words in the reference caption first and add new words from scratch to output caption, \ie, ES is calculated as the sum of words in reference and output caption. Meanwhile, we mainly focused on the efficiency evaluation of ECE models, so we have used gray font for efficiency evaluation of ICE methods. Results on COCO-EE and Flickr30K-EE are reported in Table~\ref{tab:cocoee} and Table~\ref{tab:Flickr30Kee}, respectively.


\noindent\textbf{Results on COCO-EE.} From Table~\ref{tab:cocoee}, we can observe: 1) For the quality evaluation, our model achieves the largest performance gains on all metrics (\eg, 194.8 vs. 190.5 in ETN on CIDEr-D). 2) For efficiency evaluation, SOTA implicit models always outperform their explicit counterparts, but they require more editing steps. Instead, our model achieves the best GPS(C) score by predicting more \texttt{ADD} operations, instead of simply deleting or keeping the words in the reference captions. It also shows our ability to detect and fix detailed errors.

\noindent\textbf{Results on Flickr30K-EE.} From Table~\ref{tab:Flickr30Kee}, we can observe: 1) For the quality evaluation, similar with COCO-EE, our model achieves the largest performance gains on all metrics (\eg, 148.3 vs. 143.3 in ETN on CIDEr-D). 2) For efficiency evaluation, our model achieves the best GPS(C) score (\eg, 8.58 vs. 6.90 in V-EditNTS). Compared to the weaknesses of implicit models (need more editing steps) and explicit models (marginal performance gains), our model achieves a decent balance between performance gains and editing steps, \ie, we improved the quality of reference captions with quite a few meaningful editing steps.

\begin{table}[t]
  \setlength{\abovecaptionskip}{-0.5em}
  \setlength{\belowcaptionskip}{-0.5em}
  \begin{center}
    \scalebox{0.95}{
        \begin{tabular}{c|l|ccccccc|cc|cc}
        \hline
        & \multirow{2}{*}{Model} & \multicolumn{7}{c|}{Quality evaluation} &
        \multicolumn{4}{c}{Efficiency evaluation} \\
        & & {B-1} & {B-2} & {B-3} & {B-4} & {R} & {C} & {S} & 
        {ES} & {GPS(C)} & \textcolor{gray}{D} & \textcolor{gray}{A}\\
        \hline
        & Ref-Cap & 34.7 & 24.0 & 16.8 & 10.9 & 36.9 & 91.3 & 23.4 
        & {---} & \cellcolor{mygray-bg}{---} & {---} & {---}\\
        & UpDn~\cite{anderson2018bottom} & 25.6 & 16.1 & 10.4 & 6.3 & 30.1 & 71.0 & 21.4 & {---} & \cellcolor{mygray-bg}{---} & {---} & {---}\\
        \hline
        \multirow{3}{*}{ICE} & UpDn-E~\cite{anderson2018bottom} & 33.9 & 24.7 & 18.3 & 12.5 & 41.1 & 129.1 & 29.8 & \textcolor{gray}{12.00}  & \cellcolor{mygray-bg}{\textcolor{gray}{3.15}} & \textcolor{gray}{7.41}  & \textcolor{gray}{4.59}  \\
        & MN~\cite{sammani2019look} & 30.0 & 20.0 & 13.6 & 8.6 & 34.9 & 91.1 & 25.2 & \textcolor{gray}{12.09}  & \cellcolor{mygray-bg}{\textcolor{gray}{-0.02}} & \textcolor{gray}{7.41}  & \textcolor{gray}{4.69}  \\
        & ETN~\cite{sammani2020show} & 34.8 & 25.9 & 19.6 & 13.7 & 41.8 & 143.3 & 31.3 & \textcolor{gray}{12.06}  & \cellcolor{mygray-bg}{\textcolor{gray}{4.31}} & \textcolor{gray}{7.41}  & \textcolor{gray}{4.65}  \\
        \hline
        \multirow{4}{*}{ECE} & V-EditNTS~\cite{dong2019editnts} & 38.0 & 27.6 & 20.1 & 13.8 & 40.2 & 129.1 & 28.7 & 5.48  & \cellcolor{mygray-bg}{6.90} & \textcolor{gray}{3.59}  & \textcolor{gray}{1.89}  \\
        & V-Felix~\cite{mallinson2020felix} & 21.1 & 16.7 & 13.5 & 10.1 & 38.0 & 127.4 & 27.8 & 5.54  & \cellcolor{mygray-bg}{6.51} & \textcolor{gray}{4.92}  & \textcolor{gray}{0.62}  \\
        & V-LaserTagger~\cite{malmi2019encode} & 30.8 & 20.8 & 15.0 & 10.5 & 34.9 & 104.0 & 27.3 & 3.37  & \cellcolor{mygray-bg}{3.77} & \textcolor{gray}{3.35}  & \textcolor{gray}{0.02}  \\
        \cline{2-13}
        & \textbf{TIger (Ours)} & \textbf{38.3} & \textbf{28.1} & \textbf{21.1} & \textbf{14.9} & \textbf{42.7} & \textbf{148.3} & \textbf{32.0} & 6.65 & \cellcolor{mygray-bg}{\textbf{8.58}} & \textcolor{gray}{4.63}  & \textcolor{gray}{2.02}  \\
        \hline
    \end{tabular}%
    }
  \end{center}
  \caption{Performance of our model and other state-of-art models on Flickr30K-EE. ``Ref-Caps" denotes the quality of given reference captions. ``D" and ``A" denotes the number of editing step of \texttt{DELETE} and \texttt{ADD} operations, respectively.}
  \label{tab:Flickr30Kee}%
\end{table}%

\subsection{Ablation Studies} \label{sec:5.3}

In this section, we run a set of ablation studies to analyze the influence of different hyperparameter settings, and the influence of pre-trained ViLBERT weights.

\noindent\textbf{Influence of Weighted XE Loss.}
As mentioned in Sec.~\ref{sec:4.3}, we used weighted XE loss for training. To explore the influence of different loss weights, we first run ablations by setting different loss weight ratios $\lambda \in \{1.0, 1.2, 1.5, 2.0\}$ on both Tagger$_{\text{del}}$ and Tagger$_{\text{add}}$. Results are reported in Table~\ref{tab:ablation_weight}. Then, we explored the influence of weighted XE loss to a single Tagger module, \ie, we run ablations by setting one of the Tagger with $\lambda > 1.0$, and the other with $\lambda = 1.0$. The results are reported in Table~\ref{tab:ablation_contribution_component}. Note that all Inserters were trained with $\lambda = 1.0$.

\begin{table}[t]
	\setlength{\belowcaptionskip}{-1em}

    \begin{minipage}[t]{0.4\textwidth}
    \makeatletter\def\@captype{table}
        \centering
        \scalebox{0.95}{
        \begin{tabular}{l| c|ccccc}
        \hline
        & {$\lambda$} & {B-1} & {B-4} & {R} & {C} & {S} \\
        \hline
        \parbox[t]{4mm}{\multirow{4}{*}{\rotatebox[origin=c]{90}{COCO}}} & 1.0 & 54.1 & 24.0 & 53.9 & 190.0 & 33.4 \\
        & 1.2 & 54.4 & 24.1 & 54.0 & 190.9 & 33.4 \\
        & 1.5 & \textbf{54.8} & \textbf{24.7} & \textbf{54.3} & \textbf{194.8} & \textbf{33.3} \\
        & 2.0 & 54.6 & 24.6 & 54.1 & 193.9 & 33.1 \\
        \hline
        \parbox[t]{4mm}{\multirow{4}{*}{\rotatebox[origin=c]{90}{Flickr30K}}} & 1.0 & 34.2 & 13.4 & 41.2 & 137.0 & 30.9 \\
        & 1.2 & 34.3 & 14.1 & 41.7 & 144.0 & 31.4 \\
        & 1.5 & \textbf{38.3} & \textbf{14.9} & \textbf{42.7} & \textbf{148.3} & \textbf{32.0} \\
        & 2.0 & 37.2 & 14.9 & 42.7 & 148.0 & 31.5 \\
        \hline
        \end{tabular}%
        }
       \caption{Performance on COCO-EE and Flickr30K-EE with different XE loss weights $\lambda$.}
      \label{tab:ablation_weight}%
    \label{sample-table}
    \end{minipage}
    \hspace{1.0em}
    \begin{minipage}[t]{0.55\textwidth}
    \makeatletter\def\@captype{table}
      \centering
      \scalebox{0.95}{
        \begin{tabular}{l|c c|ccccc}
        \hline
        & T$_\text{del}$ & T$_\text{add}$ & {B-1} & {B-4} & {R} & {C} & {S} \\
        \hline
        \parbox[t]{4mm}{\multirow{4}{*}{\rotatebox[origin=c]{90}{COCO}}} & & & 54.1 & 24.0  & 53.9  & 190.0  & 33.4  \\
        & \boldcheckmark &  & 55.0 & 24.7  & 54.3  & 193.7  & 33.1  \\
        &  & \boldcheckmark & 54.1 & 24.1  & 54.0  & 191.2  & 33.7  \\
        & \boldcheckmark & \boldcheckmark & \textbf{54.8} & \textbf{24.7} & \textbf{54.3} & \textbf{194.8} & \textbf{33.3} \\
        \hline
        \parbox[t]{4mm}{\multirow{4}{*}{\rotatebox[origin=c]{90}{Flickr30K}}} & & & 34.2 & 13.4  & 41.2  & 137.0  & 30.9  \\
        & \boldcheckmark &  & 34.9 & 14.3  & 42.0  & 144.9  & 34.6  \\
        & & \boldcheckmark & 34.3 & 13.7  & 41.4  & 140.9  & 31.2  \\
        & \boldcheckmark & \boldcheckmark & \textbf{38.3} & \textbf{14.9} & \textbf{42.7} & \textbf{148.3} & \textbf{32.0} \\
        \hline
        \end{tabular}%
       }
      \label{tab:ablation_contribution_component}%
      \caption{Influence of different modules with weighted XE loss ($\lambda=1.5$). ``T$_{\text{del}}$" and ``T$_{\text{add}}$" denote Tagger$_{\text{del}}$ and Tagger$_{\text{add}}$, respectively.}
    \end{minipage}
\end{table}

\noindent\textbf{Results.} From Table.~\ref{tab:ablation_weight}, we have several observations: 1) For both the COCO-EE and Flickr30K-EE, \texttt{TIger} with weighted XE loss training always gets better performance than the baseline ($\lambda = 1.0$). 2) The model trained with $\lambda = 1.5$ gets the best performance, \ie, it boosts the CIDEr-D score from 190.0 to 194.8 for COCO-EE and from 137.0 to 148.3 for Flickr30K-EE. This demonstrates the effectiveness of paying more attention to predicting the \texttt{KEEP} operation. We then used $\lambda=1.5$ to train \texttt{TIger} in all experiments. From Table~\ref{tab:ablation_contribution_component}, we can observe that: 1) \texttt{TIger} with only one of the Tagger modules trained with $\lambda=1.5$ alone can still achieve better performance than baseline. 2) The weighted XE loss has more impact on Tagger$_{\text{del}}$ than Tagger$_{\text{add}}$. The possible reason is that \texttt{TIger} only applies Tagger$_{\text{del}}$ once, which determines the basic caption for further adding.

\noindent\textbf{Different Editing Rounds.}
Since \texttt{TIger} iteratively use Tagger$_{\text{add}}$ and Inserter multiple rounds to add words, we run ablations to analyse the effect of different edit rounds. The maximum number of editing rounds was set to 5. 

\noindent\textbf{Results.}
From Table~\ref{tab:ablation_edittime}, we can observe that: 1) For COCO-EE, the performance of \texttt{TIger} keeps improving in the first 3 editing rounds. Then, the quality evaluation metrics reach the best scores and keep unchanged or even slightly drop with more editing rounds. For example, BLEU-1 keeps increasing with more editing rounds, CIDEr-D reaches the best score 194.8 in the 4-th round and drops to 194.6 in the 5-th round. Since most metrics reach their best scores in the 4-th round, considering the trade-off between model performance and editing efficiency, we used 4 editing rounds for the COCO-EE. 2) For Flickr30K-EE, the performance of \texttt{TIger} keeps improving in the first 3 editing rounds. Most quality evaluation metrics reach the best score in the 3-rd round, and then keep unchanged (14.9 for BLEU-4 and 148.3 for CIDEr-D) or drop slightly (SPICE) with more editing rounds. Thus, we used 3 editing rounds for the Flickr30K-EE.


\begin{figure}[t]
	\setlength{\abovecaptionskip}{-1em}
	\setlength{\belowcaptionskip}{-0.5em}
	\begin{center}
    	\includegraphics[width=\linewidth]{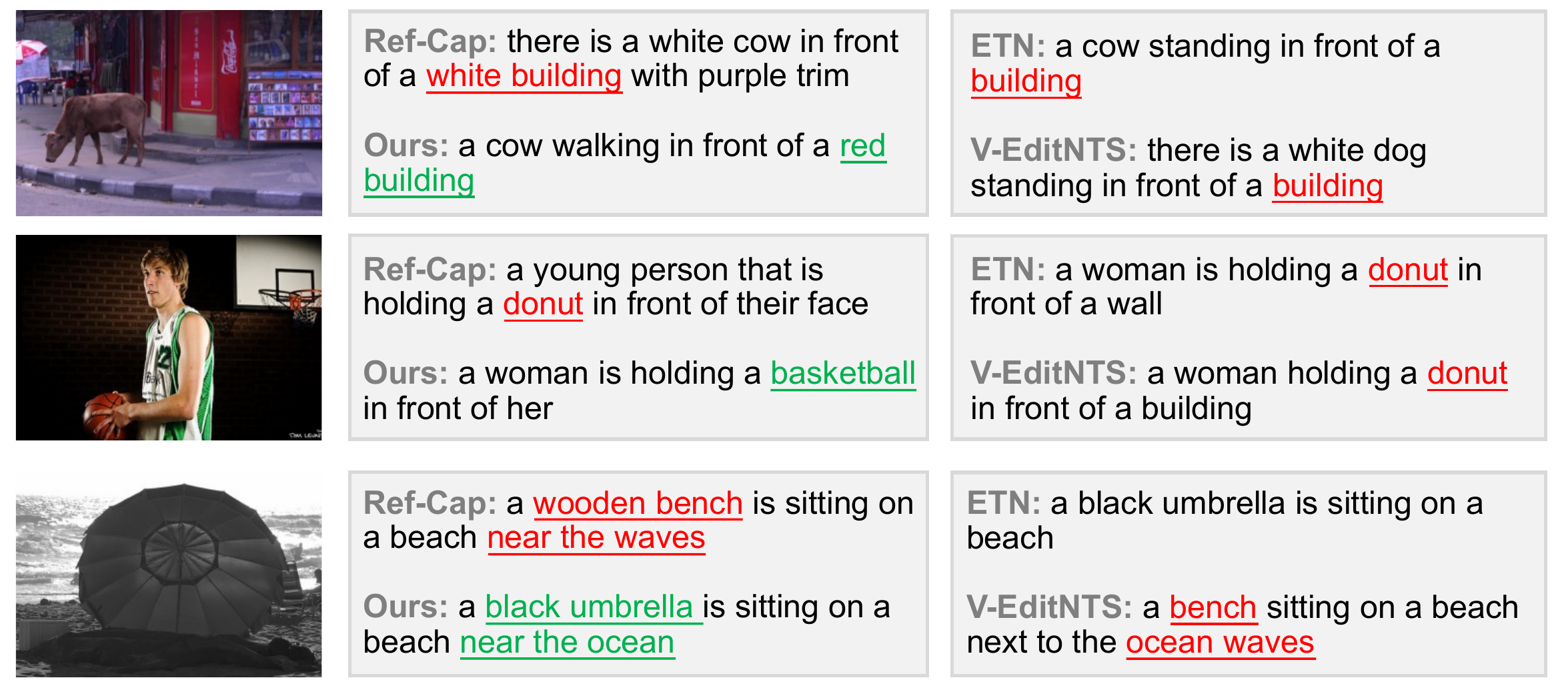}
	\end{center}
	\caption{Visualization results of our model compared to baselines in COCO-EE.}
	\label{fig:more_instance}
\end{figure}

\begin{table}[t]
  \setlength{\abovecaptionskip}{-0.5em}
  \setlength{\belowcaptionskip}{-0.5em}
  \begin{center}
    \scalebox{0.95}{
    \begin{tabular}{c|cccccc|cccccc}
    \hline 
    \multirow{2}{*}{\# Rounds} & \multicolumn{6}{c|}{COCO-EE} & \multicolumn{6}{c}{Flickr30K-EE} \\
    & {B-1} & {B-4} & {R} & {C} & {S} & {GPS(C)} & {B-1} & {B-4} & {R} & {C} & {S} & {GPS(C)}\\
    \hline
    1 & 49.3 & 22.2 & 54.2 & 180.2 & 31.6 & 8.01 & 33.5 & 13.9 & 42.2 & 142.8 & 31.0 & 8.79\\
    2 & 52.8 & 23.8 & 54.3 & 189.8 & 32.7 & 8.41 & 36.8 & 14.8 & 42.5 & 148.1 & 31.9 & \textbf{8.88}\\
    3 & 54.2 & 24.5 & \textbf{54.4} & 193.9 & 33.1 & \textbf{8.49} & 38.3 & \textbf{14.9} & 42.7 & \textbf{148.3} & \textbf{32.0} & 8.58\\
    4 & 54.8 & \textbf{24.7} & 54.3 & \textbf{194.8} & \textbf{33.3} & 8.38 & 38.4 & 14.9 & \textbf{42.8} & 148.3 & 31.9 & 8.45\\
    5 & \textbf{55.0} & 24.7 & 54.3 & 194.6 & 33.3 & 8.26 & \textbf{38.6} & 14.9 & 42.8 & 148.3 & 31.9 & 8.42\\
    \hline
    \end{tabular}%
    }
  \end{center}
  \caption{Performance of \texttt{TIger} with different editing rounds.}
  \label{tab:ablation_edittime}%
\end{table}%

\noindent\textbf{Influence of the pre-trained Weights.}
Since we took advantage of the pre-trained weights to train \texttt{TIger}, we further ran ablations to examine the influence of the pre-trained ViLBERT weights. The results are reported in Table~\ref{tab:ablation_pretrain}.

\noindent\textbf{Results.} 
From Table~\ref{tab:ablation_pretrain}, we can observe that for both datasets, as the first ECE model, both TIger models with and w/o pre-trained weights all outperform other ECE baselines with same pre-trained weights in both quality and efficiency evaluation. Meanwhile, \texttt{TIger} trained with pretraind weight achieves better performance than the model trained from scratch. For example, in CIDEr-D, the pre-trained weights improve score from 178.1 to 194.8 for COCO-EE.

\begin{table}[t]
  \setlength{\abovecaptionskip}{-0.5em}
  \setlength{\belowcaptionskip}{-0.5em}
  \begin{center}
    \scalebox{1.0}{
    \begin{tabular}{l|ccccc|ccccc}
    \hline
    \multirow{2}{*}{Models} & \multicolumn{5}{c|}{COCO-EE} & \multicolumn{5}{c}{Flickr30K-EE}\\
    & {B-1} & {B-4} & {R} & {C} & {S} & {B-1} & {B-4} & {R} & {C} & {S}\\
    \hline
    \texttt{TIger} w/o pretrain & 53.6 & 23.3  & 52.8  & 178.1  & 31.1 & 35.0 & 14.0  & 41.7  & 140.8  & 30.9 \\
    \texttt{TIger} & \textbf{54.8} & \textbf{24.7} & \textbf{54.3} & \textbf{194.8} & \textbf{33.3} & \textbf{38.3} & \textbf{14.9} & \textbf{42.7} & \textbf{148.3} & \textbf{32.0}\\
    \hline
    \end{tabular}%
    }
  \end{center}
  \caption{The Influence of the pretraind ViLBERT weight.}
  \label{tab:ablation_pretrain}%
\end{table}%

\begin{table}[t]
  \setlength{\abovecaptionskip}{-0.5em}
  \setlength{\belowcaptionskip}{-1.0em}
  \begin{center}
        \begin{tabular}{l|ccccc|ccccc}
        \hline 
        \multirow{2}{*}{Models} & \multicolumn{5}{c|}{COCO-EE} & \multicolumn{5}{c}{Flickr30K-EE} \\
         & {B-1} & {B-4} & {R} & {C} & {S} & {B-1} & {B-4} & {R} & {C} & {S}\\
        \hline
        V-EditNTS & 49.2 & 20.5 & 49.8 & 149.0 & 26.2 & 38.0 & 13.8 & 40.2 & 129.1 & 28.7\\
        \textbf{V-EditNTS+Ours} & \textbf{51.9} & \textbf{21.6} & \textbf{51.7} & \textbf{172.7} & \textbf{32.3} & {36.2} & {13.6} & \textbf{40.9} & \textbf{135.8} & \textbf{30.3}\\
        \hline
        V-Felix & 36.9 & 16.2 & 49.7 & 139.5 & 25.3 & 21.1 & 10.1 & 38 & 127.4 & 27.8\\
        \textbf{V-Felix+Ours} & \textbf{51.2} & \textbf{21.7} & \textbf{51.9} & \textbf{175.3} & \textbf{32.3} & \textbf{30.2} & \textbf{12.8} & \textbf{39.6} & \textbf{133.8} & \textbf{29.5}\\
        \hline
        V-LaserTagger & 42.0 & 16.0 & 46.8 & 127.1 & 24.1 & 30.8 & 10.5 & 34.9 & 104.0 & 27.3\\
        \textbf{V-LaserTagger+Ours} & \textbf{50.7} & \textbf{20.4} & \textbf{50.9} & \textbf{166.4} & \textbf{31.7} & \textbf{32.4} & \textbf{10.9} & \textbf{36.7} & \textbf{110.4} & {27.2}\\
        \hline
      \end{tabular}%
  \end{center}
  \caption{The result of extending \texttt{TIger} for ECE baselines}
  \label{tab:extending}%
\end{table}%

\subsection{Transfering to Machine-Generated Captions}

As mentioned before, machine-generated captions may be semantic coherent but suffer from severe bias issues, such as overlooking some content details and producing incorrect or repetitive content. To evaluate the generalization ability on machine-generated captions, we directly use the trained \texttt{TIger} to edit machine-generated captions without extra fine-tuning. To guarantee fairness and avoid data leakage, we first trained \texttt{TIger} and the ECE baselines on the same COCO-EE (and Flickr30K-EE) training set. Then, we apply the trained \texttt{TIger} to directly edit the captions generated from these ECE baselines (\ie, as reference captions) on the test set. The results are reported in Table.~\ref{tab:extending}.

\noindent\textbf{Results.} As shown in Tabel.~\ref{tab:extending}, we can observe that: 1) For COCO-EE, our proposed \texttt{TIger} can significantly improve the quality of all the captions generated by ECE baselines (\eg, CIDEr-D score from 149.0 to 172.7 for V-EditNTS). 2) For Flickr30K-EE, the average improvements are still remarkable (\eg, 135.8 vs. 129.1 in V-EditNTS on CIDEr-D score).
This also demonstrates the robustness of \texttt{TIger} when given different reference captions (\eg, these ECE baselines generated captions may erroneously delete or preserve some words).

\subsection{Qualitative Evaluation}
Fig.~\ref{fig:more_instance} shows some results generated by \texttt{TIger} compared to baselines (ETN~\cite{sammani2020show} and V-EditNTS~\cite{dong2019editnts}). The three examples demonstrate that our model is capable of recognizing and correcting incorrect details (\ie, ``\texttt{white}" to ``\texttt{red}", ``\texttt{donut}" to ``\texttt{basketball}", and ``\texttt{bench}" to ``\texttt{umbrella}"), while the baselines simply delete the wrong word ``\texttt{white}" or fail to correct the object errors ``\texttt{donut}" and `\texttt{umbrella}". Meanwhile, the last example demonstrates that our model can add new details to the captions, like attributes (color) of main objects (\eg, \texttt{black}), while baselines may overlook them. Furthermore, our model can fix these details without breaking the structure of the caption (\eg, \texttt{near the ocean}).

\section{Conclusions and Future Work}
In this paper, we proposed a new visual-language task: Explicit Caption Editing (ECE). To facilitate the ECE research, we also proposed two benchmarks by re-organizing two existing datasets MSCOCO and e-SNLI-VE, dubbed as COCO-EE and Flickr30K-EE, respectively. Meanwhile, we proposed the first ECE model \texttt{TIger}. We validate the effectiveness of \texttt{TIger} through extensive comparative and ablative experiments. Moving forward, we are going to 1) design stronger ECE models by introducing some advanced edit operations; 2) try to bridge the gap between explicit and implicit editing, and propose a unified model for both tasks.

\begin{footnotesize}
\textbf{Acknowledgement.}
This work was supported by the National Key Research \& Development Project of China (2021ZD0110700), the National Natural Science Foundation of China (U19B2043, 61976185), Zhejiang Natural Science Foundation (LR19F020002), Zhejiang Innovation Foundation(2019R52002), and the Fundamental Research Funds for the Central Universities (226-2022-00051).
\end{footnotesize}

%
%
\bibliographystyle{splncs04}
\bibliography{egbib}

\clearpage

\appendix

\begin{center}%
  {\LARGE \textbf{*** Supplementary Manuscript ***} \par}%
\end{center}

\vskip 3em

The supplementary manuscript is organized as follows:

\begin{enumerate}[leftmargin=1cm]

    \item[$\bullet$] In Sec.~\ref{sec:a}, we provide more detailed construction steps for both COCO-EE and Flickr30K-EE (\cf~Sec.~\ref{sec:3.2}).
    
    \item[$\bullet$] In Sec.~\ref{sec:b}, we discuss the two ways of adding new words in Tagger$_\text{add}$.
    
    \item[$\bullet$] In Sec.~\ref{sec:c}, we explain more details about the calculation of the proposed metric Editing Steps (ES)  (\cf~Sec.~\ref{sec:5.1}). 
    
    \item[$\bullet$] In Sec.~\ref{sec:d}, we show the implementation details.
    
    \item[$\bullet$] In Sec.~\ref{sec:e}, we show the details and architectures of the compared baselines (\cf~Sec.~\ref{sec:5.2}).
    
    
    \item[$\bullet$] In Sec.~\ref{sec:f}, we show the computational efficiency of \texttt{TIger} and the compared ECE baselines.
    
    \item[$\bullet$] In Sec.~\ref{sec:g}, we illustrate more visualization results generated by \texttt{TIger}.

\end{enumerate}

\section{More Detailed Benchmark Construction Steps}
\label{sec:a}

\subsection{COCO-EE}
We built COCO-EE based on MSCOCO~\cite{lin2014microsoft}, which contains 123,287 images, and 5 ground-truth captions for each image. To ensure \emph{criteria~1}, we selected all Ref-Caps and GT-Caps in COCO-EE from MSCOCO captions. Specifically, we constructed each editing instance following these steps:
\begin{enumerate}
    \item \textbf{Image-Caption Similarity Filter.}  To guarantee \emph{criteria~2}, for each image labeled with 5 captions,  we used a pre-trained CLIP~\cite{radford2021learning} model to filter 300 captions from all the rest captions in its respective split set ( training/validation/test) based on the CLIP score, where a higher score indicates higher similarity between the image and the caption. Meanwhile, to save the computation cost in the rest filtering steps, we then randomly selected 30 captions from them as Ref-Cap candidates, and we treated all the 5 ground-truth captions as the GT-Cap candidates.
    
    \item \textbf{Caption Similarity Filter.} To guarantee \emph{criteria~3}, we filtered each image's Ref-Cap candidates based on the BLEU~\cite{papineni2002bleu} score between the Ref-Cap and GT-Cap candidates. We only kept the Ref-Cap candidates whose BLEU scores are greater than a certain threshold ( BLEU-2~\textgreater~0.4 \& BLEU-3~\textgreater~0.3).
    
    \item \textbf{Caption Differences Filter.} To guarantee \emph{criteria~4}, we filtered each image's Ref-Cap candidates based on the SPICE~\cite{anderson2016spice} score between the Ref-Cap and GT-Cap candidates. The SPICE scores reflect the similarity of scenes described by different captions, and we only kept the Ref-Cap candidate whose SPICE score is less than a certain threshold (\ie, SPICE~\textless~0.35 ). 
    
    \item \textbf{Edit Distance Filter.} Finally, for each filtered Ref-Cap candidate, we only selected the caption with the shortest edit distance from the corresponding GT-Cap candidates to form a Ref-GT caption pair. 
    
\end{enumerate}

Following the above steps, we constructed the COCO-EE, and divided it into training, validation, and test sets following the “Karpathy” split~\cite{karpathy2015deep}.

\begin{figure*}[t]
\centering
\includegraphics[width=0.8\linewidth]{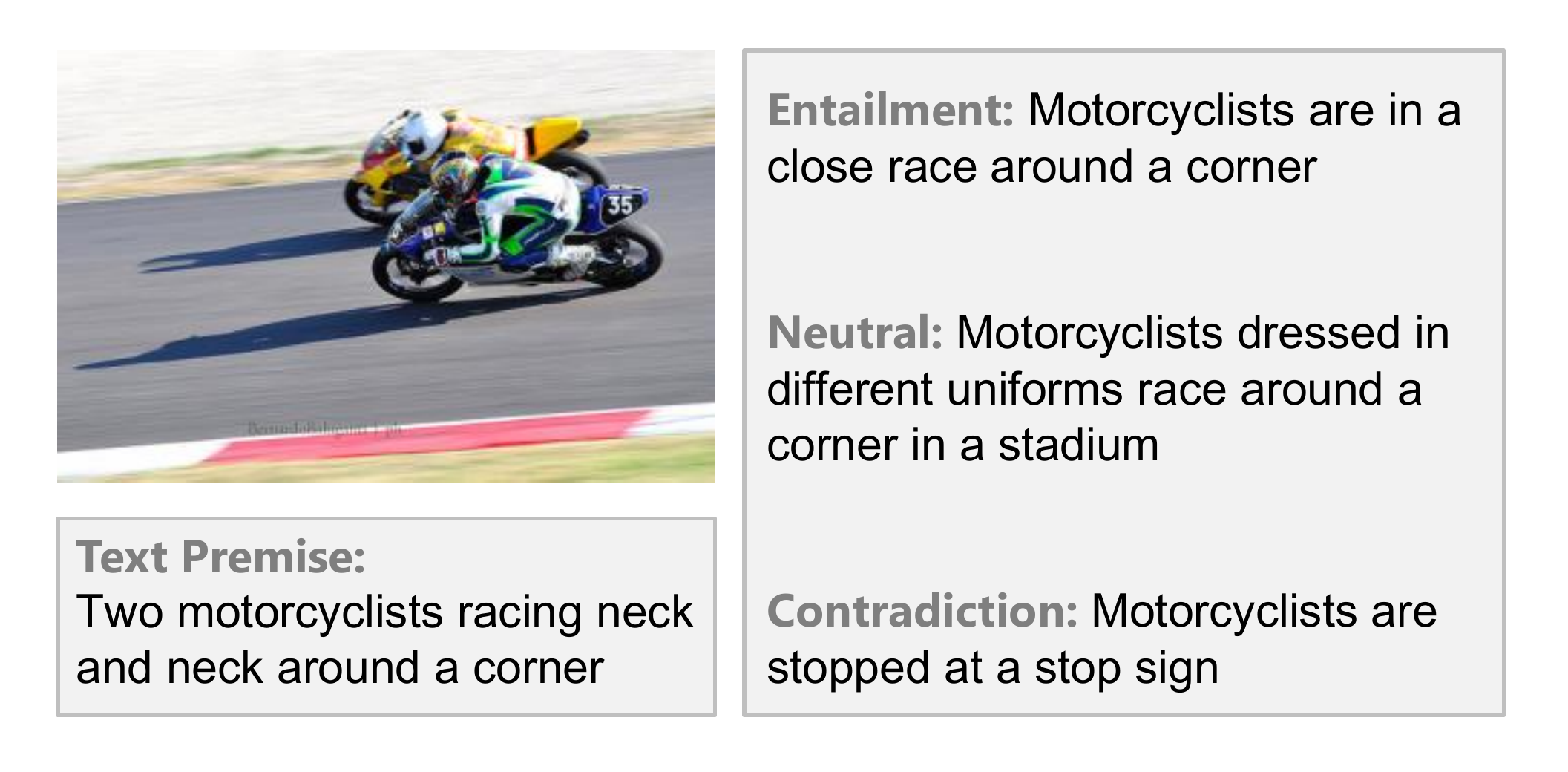} 
\caption{Instance from e-SNLI-VE.}
\label{fig:instance_esnlive}
\end{figure*}

\subsection{Flickr30K-EE} 

We built Flickr30K-EE based on dataset e-SNLI-VE~\cite{kayser2021vil}. e-SNLI-VE is a visual entailment dataset using the same image set as the image captioning dataset Flicrk30K~\cite{young2014image}. Specifically, e-SNLI-VE was built based on the text entailment SNLI~\cite{bowman2015large} dataset, SNLI used the captions from Flickr30K as text premises. For each text premise, there are three human-annotated sentence hypotheses, and each sentence hypothesis has a different relationship with the text premise. The e-SNLI-VE then replaced all the text premises with corresponding Flickr30K images and relabeled the sentence hypothesis to correct labeling errors.

Fig.~\ref{fig:instance_esnlive} shows an instance of e-SNLI-VE. For each image in e-SNLI-VE, there are three sentence hypotheses, and each sentence hypothesis has a different relationship with the image premise: 

\begin{enumerate}[leftmargin=1cm]

    \item[$\bullet$] \textbf{Entailment}: if there is enough evidence in the image premise to conclude that hypothesis is true.
    
    \item[$\bullet$] \textbf{Neutral}: if there is not enough evidence to conclude whether the hypothesis is true or false.
    
    \item[$\bullet$] \textbf{Contradiction}: if there is enough evidence in the image premise to conclude that hypothesis is false.

\end{enumerate}

For each image and its textual hypotheses in the e-SNLI-VE, we only selected the contradiction and entailment hypothesis as a Ref-GT caption pair if they used to have the same text premise. We divided it into training, validation, and test sets based on e-SNLI-VE splits.

\section{Two Ways of Adding New Words in Tagger$_\text{add}$}
\label{sec:b}

\begin{figure*}[t]
\centering
\includegraphics[width=\linewidth]{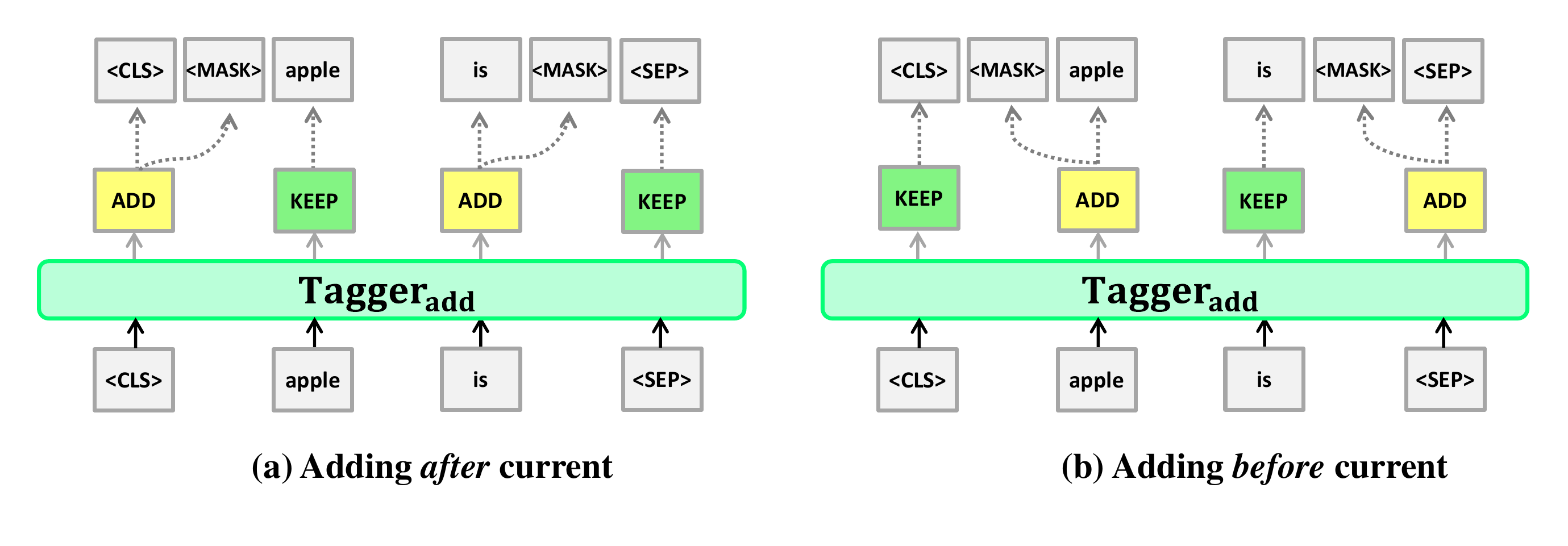} 
\vspace{-2.5em}
\caption{Two ways of adding new words in Tagger$_{\text{add}}$.}
  \vspace{-1.0em}
\label{fig:difference_adding}
\end{figure*}

As shown in Fig.~\ref{fig:difference_adding}, there are two ways to add new words in Tagger$_\text{add}$. Take the simple caption ``\texttt{apple is}" as an example, to change it into ``\texttt{the apple is red}", each newly added word (\ie, [\texttt{MASK}] token) can be added either \emph{after} the current token or \emph{before} the current token. 

When adding \emph{after} the current token, [\texttt{CLS}] token should predict \texttt{ADD}, meanwhile we never need to add after the final [\texttt{SEP}] token, \ie, the ground-truth edit operation for [\texttt{SEP}] token is constant (\texttt{KEEP}) and could be excluded from the loss computations. When adding \emph{before} the current token, [\texttt{SEP}] token should predict \texttt{ADD}, and [\texttt{CLS}] is constant to predict \texttt{KEEP} because we never need to add words before it. There is no essential difference between the two ways in terms of model training. In our experiments, we use the way of adding \emph{after} the current token for \texttt{Tiger}.

\section{Details of Calculating the Editing Steps}
\label{sec:c}

As mentioned in Sec.~\ref{sec:5.1}, the metric Editing Steps (\textbf{ES}) is the total number of meaningful editing steps, in this paper, we regard the sum of \texttt{DELETE} and \texttt{ADD} operations as \textbf{ES} since all baselines apply the same set of edit operations (\ie, the three basic operations \texttt{KEEP}, \texttt{DELETE}, \texttt{ADD}, and the combination of them). 

Specifically, if an edit operation is a combination of the above basic operations, we can decompose them into the three basic operations. For the V-Felix and V-LaserTagger baselines, we calculated ES as follows:

\begin{enumerate}
    \item Felix~\cite{mallinson2020felix} predicts the number of adding words together with the keep and delete operation, \eg, the edit operation \texttt{(DELETE|N)} means delete the current token and add N new words after this token. This can save the number edit operation compared to predict each \texttt{DELETE} and \texttt{ADD} operation individually, but their contribution to the editing process are the same, and \texttt{(DELETE|N)} essentially achieves the editing by each \texttt{DELETE} and \texttt{ADD} operation separately. We thus count \texttt{(DELETE|N)} as one \texttt{DELETE} operation and N \texttt{ADD} operations (N+1 editing steps). Similarly, \texttt{(KEEP|N)} is counted as one \texttt{KEEP} operation and N \texttt{ADD} operations (N editing steps).
    
    \item LaserTagger~\cite{malmi2019encode} predicts new words before the deleted or preserved tokens, \eg, the edit operation \texttt{(this is|DELETE)} means delete the current token and add two new words ``\texttt{this is}" before this token. Thus, we count it as one \texttt{DELETE} operation and two \texttt{ADD} operations for \textbf{ES} computing. Similarly, \texttt{(this is|KEEP)} is counted as one \texttt{KEEP} operation and two \texttt{ADD} operations.

\end{enumerate}

 Besides the basic edit operations and their combination operations, different (or future) ECE models may design other special or high-level edit operations. For edit operations that essentially changes the token in the sentence, we split them into basic \texttt{DELETE} and \texttt{ADD} operations for \textbf{ES} computing, \eg, \texttt{REPLACE}, which replace the current token with a new one, we regard it as deleting the current token and adding a new one, so it will be counted as one \texttt{DELETE} operation and one \texttt{ADD} operation. For other edit operations that only change the order of the input tokens, we count each of them as one editing step.

 \section{Implementation Details} \label{sec:d}
 For visual token features, we used the same bottom-up features from~\cite{anderson2018bottom}, which are extracted by a Faster R-CNN~\cite{ren2015faster} pre-trained on VG~\cite{sharma2018conceptual}. For multimodal BERTs, we used the 12-layer base ViLBERT model~\cite{lu2019vilbert} and used the checkpoint pre-trained on the Conceptual Captions~\cite{sharma2018conceptual} for initialization. All three modules are trained separately with a XE loss. The batch size was set to 64. We trained these modules with Adam optimizer for 20 epochs, and the initial learning rate was set to 2e-6. We used a linear decay learning rate schedule with warm up to train these modules. Besides, we expanded the editing instances based on the iterative editing process to train Tagger$_{\text{add}}$ and Inserter (\eg, an editing instance which needs a three-round editing will be expanded into three training samples corresponding to three rounds respectively).

\section{Details of these Compared Baselines}
\label{sec:e}

In this section, we describe the detailed architectures of the compared state-of-the-art baselines.

\begin{figure*}[t]
\centering
\includegraphics[width=\linewidth]{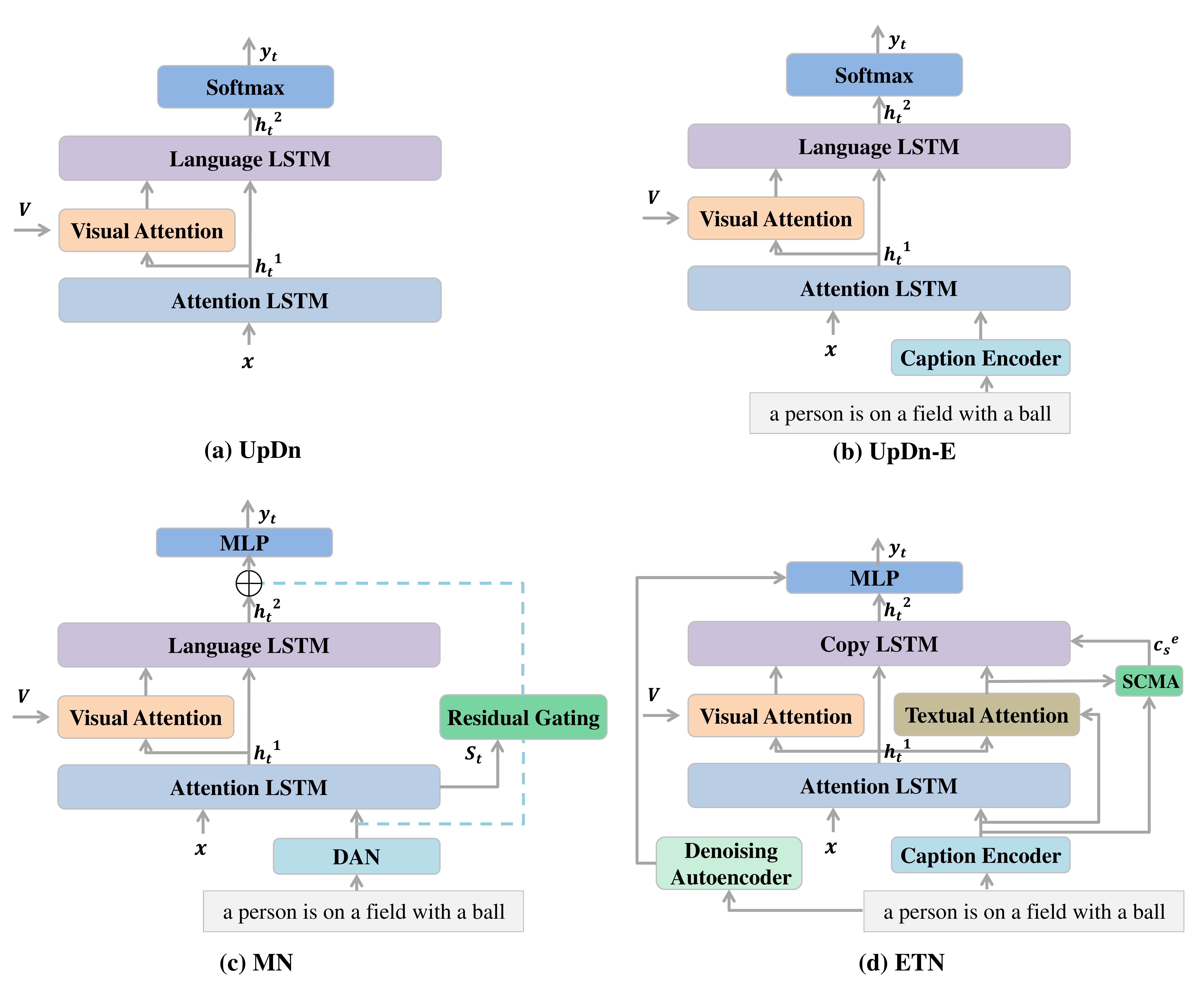}
\vspace{-2em}
\caption{Architectures of the implicit caption editing baselines}
  \vspace{-1.0em}
\label{fig:arche_imp_baselines}
\end{figure*}

Three implicit caption editing baselines are all built on top of the widely-used UpDn architecture~\cite{anderson2018bottom}. Fig.~\ref{fig:arche_imp_baselines} shows their architectures.

In UpDn, the input vector to the attention LSTM at each time step consists of the previous output of the language LSTM, concatenated with the mean-pooled image feature and the encoding of the previously generated word. The output of the language LSTM at each time step is then used to predict the output word.

\begin{enumerate}
    \item \textbf{UpDn-E}~\cite{anderson2018bottom}: It uses an extra caption encoder to encode the Ref-Cap, the caption encoder is a bi-directional LSTM same as the one in ETN~\cite{sammani2020show}, and the output of caption encoder is concatenated to the input vector; 
    \item \textbf{MN}~\cite{sammani2019look}: It uses a pre-trained Deep Averaging Network (DAN) to encode the Ref-Cap, the output of the DAN is concatenated to the input vector. A residual LSTM gate is used to extract residual information from attention LSTM and DAN, which are then summed with the output of the language LSTM to predict the output word.
    \item \textbf{ETN}~\cite{sammani2020show}: It uses a Selective Copy Memory Attention (SCMA) to select and copy memory states corresponding to words in the Ref-Cap, and a Copy-LSTM to generate words. Meanwhile, it uses a bi-directional LSTM to encode the Ref-Cap. Besides, it further uses a denoising autoencoder to boost Copy-LSTM and gets the final prediction.
\end{enumerate}

\begin{figure*}[!h]
\centering
\includegraphics[width=\linewidth]{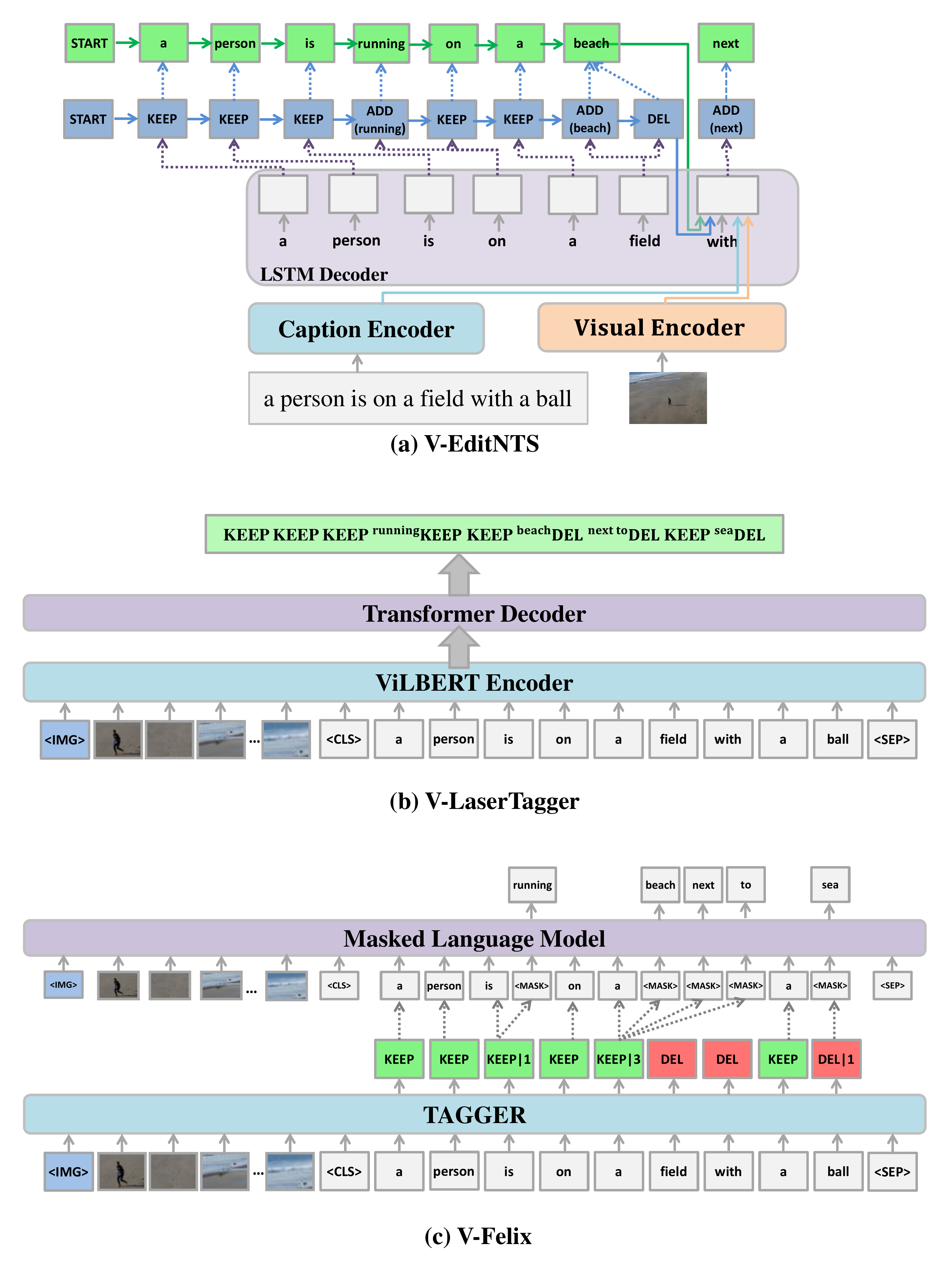}
\vspace{-3em}
\caption{Architecture of the three explicit baselines.}
 \vspace{-2em}
\label{fig:arche_exp_baselines}
\end{figure*}

We also extended three text explicit editing models into ECE. Fig.~\ref{fig:arche_exp_baselines} shows their architectures.
\begin{enumerate}
    \item \textbf{V-EditNTS}~\cite{dong2019editnts}: It predicts edit operation sequence iteratively by an LSTM, including \texttt{KEEP}, \texttt{DELETE}, and \texttt{ADD}, it also predicts the specific word for \texttt{ADD} at the same time. We used an extra visual encoder that projects the mean-pooled image feature to match the dimension of the caption encoding. The input vector to the decoder LSTM at each time step consists of the caption encoding of Ref-Cap, concatenated with the visual encoding of the input image, the embedding of the current token, the LSTM encoding of the previous edit operation and previous output word. 
    \item \textbf{V-LaserTagger}~\cite{malmi2019encode}: It combines a ViLBERT~\cite{lu2019vilbert} encoder with an autoregressive Transformer decoder. Specifically, it uses three edit operations: keeping a token, deleting a token, and adding new words before the token. The adding words are restricted to the phrase vocabulary that is derived from respective training data (COCO-EE and Flickr30K-EE)\footnote{We also tried different vocabulary sizes (\eg, fixing to 500 as in~\cite{malmi2019encode}), the empirical results are similar without obvious differences.}.
    \item \textbf{V-Felix}~\cite{mallinson2020felix}: It is composed of two ViLBERT~\cite{lu2019vilbert} models including a tagging model that predicts operations to keep or delete the token, it will also predict the number of new words to add after the token. And an insertion model that predicts specific words for new adding. For fairness, we used the V-Felix without the reordering mechanism and the way of insertion was set to [\texttt{MASK}] prediction. 
\end{enumerate}

\begin{table}[t]
  \begin{center}
    \scalebox{1}{
        \begin{tabular}{l|c|cc|cc|c}
        \hline
        \multirow{2}{*}{Model} & \multirow{2}{*}{Type} & \multicolumn{2}{c|}{COCO-EE} &
        \multicolumn{2}{c|}{Flickr30K-EE} & \multirow{2}{*}{FLOPs(M)}\\
         & & {IT(ms)} & {C} & {IT(ms)} & {C} \\
        \hline
        V-EditNTS [\textcolor{green}{9}] & L & 68.11 & 149.0 & 51.51 & 129.1 & 4.55 \\
        V-Felix [\textcolor{green}{24}] & T & 93.80 & 139.5 & 76.40 & 127.4 & 15.02 \\
        V-LaserTagger [\textcolor{green}{25}] & T & 325.45 & 127.1 & 313.08 & 104.0 & 10.42\\
        \textbf{TIger (round=1)} & T & 105.46 & 180.2 & 105.45 & 142.8 & 22.53\\
        \textbf{TIger (round=2)} & T & 172.83 & 189.8 & 170.00 & 148.1 & 37.55 \\
        \textbf{TIger (round=3)} & T & 237.93 & 193.9 & 238.62 & 148.3 & 52.57 \\
        \textbf{TIger (round=4)} & T & 304.36 & 194.8 & 302.21 & 148.3 & 67.59 \\
        \hline
        \end{tabular}%
      }
  \end{center}
  \caption{Results of average inference time (IT), CIDEr-D (C) and inference FLOPs of ECE models. Type ``L" and ``T" denote LSTM-based and Transformer-based models, respectively.}
  \label{tab:computational_efficiency}%
\end{table}%

\section{Computational Efficiency}
\label{sec:f}
For more complete experiment results, we reported general computational efficiency evaluation metrics of our model and the ECE baselines. As shown in Table~\ref{tab:computational_efficiency}, due to model structure, Transformer-based models tend to have longer inference time and larger FLOPs than LSTM-based counterparts. Indeed, the computational cost of TIger increases continuously with more round editing. However, one-round TIger can still achieve much superior performance with nearly computational efficiency compared to other ECE baselines. In this paper, we hope TIger can serve as a strong baseline, and we mainly focus on \textbf{editing efficiency} of ECE models. In contrast, \emph{the computational cost like FLOPs was hardly reported as a key metric of models in existing caption generation or editing works}.

\begin{figure}[t]
	\centering
	\includegraphics[width=\linewidth]{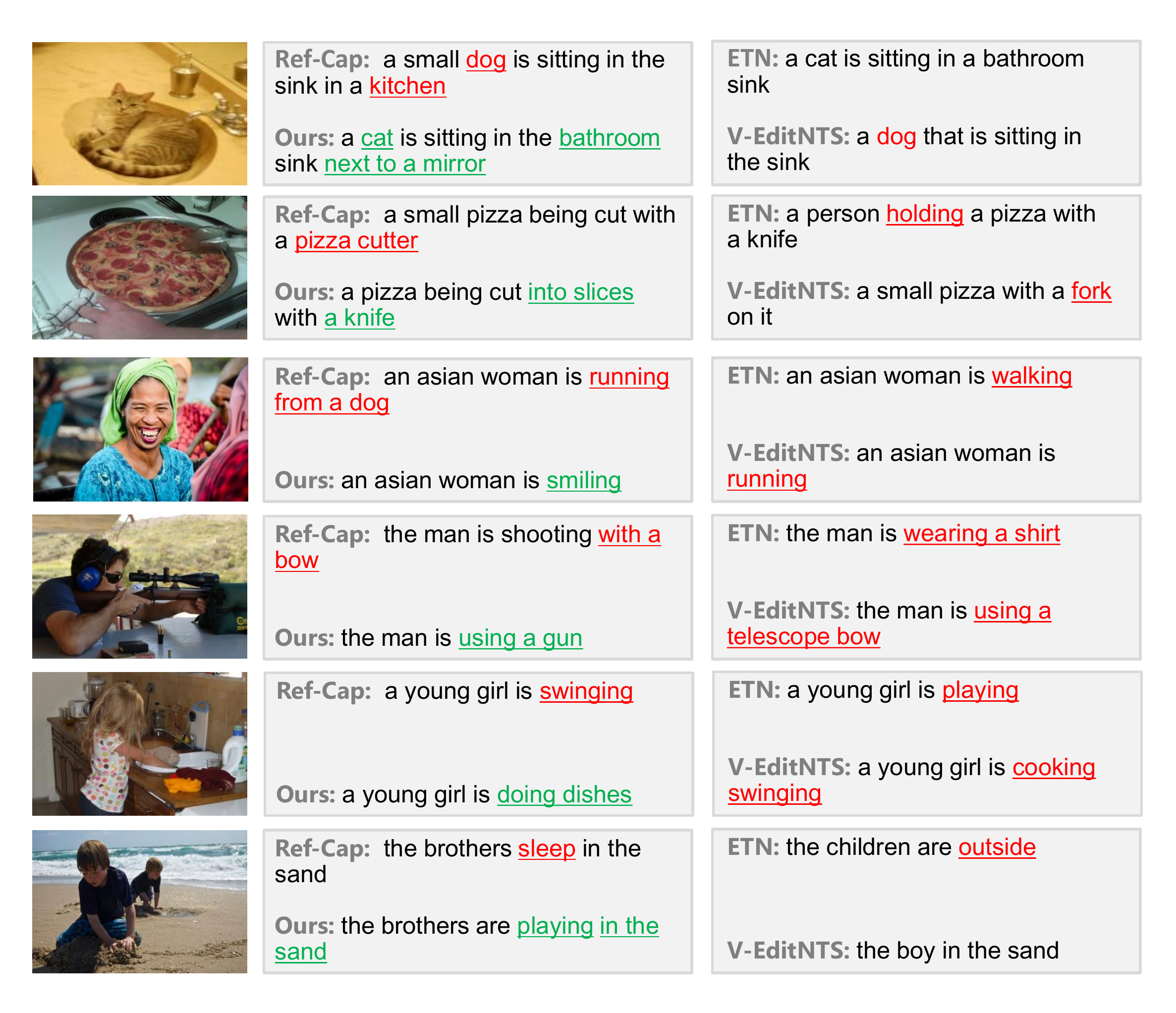}
	\vspace{-3em}
	\caption{Visualization results of our model compared to two baselines (ETN and V-EditNTS) in COCO-EE (top two samples) and Flickr30K-EE (bottom four samples).}
	\label{fig:more_instance2}
\end{figure}

\section{More Qualitative Results}
\label{sec:g}

Fig.~\ref{fig:more_instance2} shows more results generated by \texttt{TIger} compared to baselines. The first two samples are from COCO-EE, we can observe that our model is not only capable of recognizing and correcting incorrect details (\ie, ``\texttt{dog}" to ``\texttt{cat}", ``\texttt{kitchen}" to ``\texttt{bathroom}", and ``\texttt{cutter}" to ``\texttt{knife}") but also adding new details (\eg, ``\texttt{next to a mirror}"). The rest examples are from Flickr30K-EE and their Ref-Caps are relatively short, we can observe that our model can still correct the incorrect details (\eg, change from ``\texttt{running from a dog}" to ``\texttt{smiling}", ``\texttt{with a bow}" to ``\texttt{using a gun}", ``\texttt{swinging}" to ``\texttt{doing dishes}, and ``\texttt{sleep}" to ``\texttt{playing}"") without breaking the structure of the caption (\eg, \texttt{in the sand}).

%
%

\end{document}